%% file: 00-main.tex
\documentclass[10pt]{article}

\usepackage[margin=1in]{geometry} 
\usepackage[hang,flushmargin]{footmisc} 

\usepackage{fontawesome5} 

\usepackage{fancyhdr}
\usepackage{setspace}
\usepackage{sectsty}
\sectionfont{\fontsize{13}{13}\selectfont}
\subsectionfont{\fontsize{12}{12}\selectfont}

\usepackage{titlesec}
\titlespacing*{\paragraph}
    {0pt}                   
    {0.25\baselineskip}     
    {1em}                   

\usepackage{amsfonts,amssymb,amsmath,amsthm,mathtools}
\numberwithin{equation}{section}
\usepackage{bbold}

\usepackage{caption,subcaption}
\usepackage[table]{xcolor}
\definecolor{mydarkblue}{rgb}{0, 0.08, 0.45}
\usepackage{booktabs,graphicx,float,multirow,longtable}

\usepackage[ruled,algoruled,boxed]{algorithm2e}

\usepackage{enumitem}

\usepackage{natbib}
\usepackage{hyperref}
\hypersetup{colorlinks=true,linkcolor=mydarkblue,urlcolor=blue,citecolor=mydarkblue}

\input{custom-commands}

\makeatletter
\def\nomarkerfootnote{\xdef\@thefnmark{}\@footnotetext}
\makeatother

\theoremstyle{remark}
\newtheorem{remark}{Remark}
\hsize\textwidth
  \linewidth\hsize 

\title{\fontsize{16}{20}\selectfont
Deep Learning-based Approaches for State Space Models:\\
A Selective Review}
\author{
\fontsize{12}{12}\selectfont 
Jiahe Lin 
\qquad\qquad 
\fontsize{12}{12}\selectfont
George Michailidis {\scriptsize\faEnvelope[regular]}
\vspace*{0.3in}\\
\fontsize{11}{12}\selectfont {\scriptsize\faEnvelope[regular]} Department of Statistics and Data Science, UCLA
}

\date{}

\begin{document}
\maketitle
\nomarkerfootnote{{\scriptsize\faEnvelope[regular]} Corresponding author: George Michailidis   \textlangle\texttt{gmichail@ucla.edu}\textrangle} 
\nomarkerfootnote{{\scriptsize\faGithub[regular]} Code repo: \href{https://github.com/GeorgeMichailidis/unified-neural-ssm}{github.com/georgemichailidis/unified-neural-ssm}; 
it provides unified implementations of representative neural SSM backbones to facilitate model comparison, experimentation,
and practical use.}
\nomarkerfootnote{{\scriptsize\faCheckCircle[regular]} Accepted by Statistical Science}

\begin{abstract}
State-space models (SSMs) offer a powerful framework for dynamical system analysis, wherein the temporal dynamics of the system are assumed to be captured through the evolution of the latent states, which govern the values of the observations.
This paper provides a selective review of recent advancements in deep neural network-based approaches for SSMs, and presents a unified perspective for discrete-time deep state space models and continuous-time ones such as latent neural Ordinary Differential and Stochastic Differential Equations. It starts with an overview of the classical maximum likelihood based approach for learning SSMs, reviews the variational autoencoder as a general learning pipeline for neural network-based approaches in the presence of latent variables, and discusses in detail representative deep learning models that fall under the SSM framework. Very recent developments, where SSMs are used as standalone architectural modules for improving efficiency in sequence modeling, are also examined. Finally, we present examples involving nowcasting and irregularly sampled time-series data, together with a brief overview of scientific applications based on mechanistically structured SSMs.
\end{abstract}

\input{01-intro}
\input{02-intro-SSM}
\input{03-neural-SSM}
\input{04-lssl-s4-mamba}

\input{05-tasks}

\input{06-conclusion}

\vskip 0.5in
\begingroup
\small
\setlength{\bibsep}{3pt}
\bibliographystyle{chicago}
\bibliography{ref}
\endgroup

\clearpage
\appendix

\input{appendices}

\end{document}

%% file: custom-commands.tex
\DeclarePairedDelimiter\floor{\lfloor}{\rfloor}
\DeclareMathOperator*{\argmax}{argmax}
\newcommand{\KL}[2]{{\mathrm{KL}\Big(#1\,\big\|\,#2\Big)}}
\newcommand{\sKL}[2]{{\mathrm{KL}\big(#1\,\|\,#2\big)}}

\newcommand{\dd}{\text{d}}

\newcommand{\indep}{\perp\!\!\!\perp}

\newcommand{\bmu}{{\boldsymbol{\mu}}}

\newcommand{\bepsilon}{{\boldsymbol{\epsilon}}}

%% file: 01-intro.tex
\section{Introduction}

Latent space modeling is a widely used technical framework that aims to capture the essential features and patterns of complex data through low-dimensional representations, and it plays a central role in many learning tasks such as representation learning, link prediction, and dynamical system analysis.
Within the domain of dynamical system analysis, a latent space framework assumes that the dynamics of the underlying system are governed by a low-dimensional {\em latent state} process, which dictates the evolution of the observed data. By separating latent state evolution from the observation process, the resulting {\em state space models} (SSMs) provide a structured framework for modeling underlying temporal patterns and dependence in the observed data.

The state space modeling paradigm represents a well-established framework with a long and rich history. It was originally developed for the study of dynamical systems and control problems, such as predicting system behaviors and designing control strategies in engineering, where understanding the evolution of hidden states and their impact on observable outputs is crucial for system stability and optimization \citep{friedland2005control,zadeh2008linear,aoki2013state,kumar2015stochastic}. The original SSM formalism focused on systems governed by linear dynamics under a Gaussianity assumption \citep{kalman1960new,kalman1961new}. 
Over time, the framework has been extended to accommodate nonlinear dynamics and non-Gaussian models, enhancing its versatility for a wide range of real-world applications. However, the presence of nonlinearity when coupled with high-dimensionality presents significant challenges, partly due to the limitations of available technical tools. Advances in deep learning have led to renewed interest in modeling dynamical systems via SSMs; in particular, the flexible parameterization via neural networks and the variational autoencoder \citep[VAE,][]{kingma2013auto}-based learning pipeline have addressed some of the limitations of earlier methods.

The main objective of this paper is to review recent advances in deep learning techniques for SSMs, in the context of sequence modeling. Given the well-established SSM framework for dynamical systems, Section~\ref{sec:intro-SSM} provides an overview of the model formulation, including its general-form representation that encompasses the state and the observation equations. Classical strategies that leverage maximum likelihood estimation to learn SSM parameters from data are presented, and the roles of key tools such as filtering and smoothing are highlighted. As a technical preliminary, we also provide a brief review of VAE and its corresponding pipeline for modeling sequential data, which have become the standard approach for training neural network-based SSMs.

Section~\ref{sec:neural-SSM} reviews work in the literature, wherein the model formulation corresponds to an SSM with the state and observation equations parameterized with neural networks, and  employs deep learning-based approaches to handle the learning task. In the strand of deep-learning-based SSM literature reviewed here, earlier works remain relatively close to the classical framework. This is reflected either in the model formulation itself, such as assuming that latent states evolve in {\em discrete time} with fixed intervals, as reviewed in Section~\ref{sec:discrete-deep-SSM}, or in the learning methodology, which continues to rely on filtering and smoothing ideas inherited from classical SSMs; see Section~\ref{sec:early-DL-SSM}. Later work in this literature moves toward more flexible formulations, for example by allowing latent states to evolve in continuous time, and increasingly adopts VAE-based pipelines as the dominant learning strategy (Sections~\ref{sec:neuralODE} and~\ref{sec:latent-neuralSDE}). A key distinction between the two learning strategies lies in how they handle latent states. In the classical framework, filters and smoothers are employed to recursively update the distribution of latent states, using either closed-form updates or Monte Carlo approximations. In contrast, VAEs leverage an encoder and a decoder, wherein the encoder learns an approximate posterior of the latent states conditional on the observations, while the decoder reconstructs the data from latent states sampled from the former. Learning is done by maximizing a variational lower bound of the data likelihood. VAEs learn the SSM's dynamics in an end-to-end manner, by updating the neural network weights that are used for parameterizing the encoder and the decoder; they can handle more complex dependencies and relationships that are cumbersome or challenging to capture by classical filtering and smoothing techniques.

The material presented in Sections~\ref{sec:intro-SSM} and~\ref{sec:neural-SSM} focuses on SSM being a modeling framework for time series data, covering both classical approaches and deep learning based ones. 
Section~\ref{sec:semi-implicit} shifts focus and considers SSMs as modules for {\em input-output mapping} embedded within neural network architectures. This line of work, emerging since 2021, represents a stream of effort dedicated to enhancing the models' capabilities to handle long-range dependencies in sequence modeling, such as GPT-style language models \citep[e.g.,][]{radford2018improving}. In particular, compared with Transformers~\citep{vaswani2017attention}, its  ability to efficiently model dependencies across long input sequences allows for significant reductions in memory and computational requirements. For the material presented in Section~\ref{sec:semi-implicit}, the SSMs are not used as a postulated model for the data generative process, but rather they are standalone technical tools for handling sequence data.

The paper concludes by providing concrete examples that showcase that a formulation akin to the SSM form is both beneficial and convenient for selected tasks, such as modeling mixed frequency and irregularly spaced time series data. Finally, it is worth noting that this review does not cover the extensive body of work on SSMs for {\em control} problems, which constitute a separate topic on their own.

\textbf{Notation.\ } Boldface letters are used to denote multivariate random processes indexed by $t$. 
Specifically, $\boldsymbol{z}_t$ denotes the $d$-dimensional \textit{latent} (state) process and $\boldsymbol{x}_t$ the $p$-dimensional \textit{observation} process with $t$ taking values in a discrete index set $\mathcal{T}=\{0,1,2,\cdots,T\}$. In the case of a continuous index set $\mathcal{T}=[0,T]$, we use $\boldsymbol{z}(t),\boldsymbol{x}(t)$ to denote the processes and $\boldsymbol{z}_t, \boldsymbol{x}_t$ their values sampled at $t$. $\boldsymbol{u}_t$ and $\bepsilon_t$ denote $d$- and $p$-dimensional shock/noise processes associated with the state and observation processes, respectively, and $\boldsymbol{u}(t)$ and $\bepsilon(t)$ are analogously defined. The trajectory of a process from time $s$ to time $t$ is denoted by $\mathbb{x}_{s:t}:=\{\boldsymbol{x}_s,\cdots, \boldsymbol{x}_t\}$. Deterministic multivariate functions of appropriate dimensions are denoted by the lower case letters $f(\cdot), g(\cdot)$ and $h(\cdot)$, possibly indexed by $t$ if they are time-varying. $p_\theta$ is used generically for densities or probability mass functions with respect to an appropriate dominating measure. Matrices are denoted by capital letters (e.g., $A,B$). Finally, lower case Greek letters denote the parameters of appropriate dimensions of a distribution or a function; e.g., $\theta, \phi$, etc. 

%% file: 02-intro-SSM.tex
\section{Overview of State Space Models} \label{sec:intro-SSM}

SSMs are a versatile modeling framework extensively used across various scientific domains, particularly for analyzing time series data. They have a hierarchical structure encompassing two multivariate processes: (i) a $d$-dimensional {\em latent state} process $\boldsymbol{z}_t\in\mathcal{Z}$ that captures the dynamics of the evolution of the system, and (ii) a $p$-dimensional {\em observation} process $\boldsymbol{x}_t\in\mathcal{X}$ comprising the measurements, whose values are dictated by the latent state.
In this review, we focus on cases where both processes take continuous values, namely, $\mathcal{Z}\subseteq\mathbb{R}^d$ and $\mathcal{X}\subseteq\mathbb{R}^p$.\footnote{This restriction reflects the scope of the present review, rather than a fundamental boundary of the broader state-space framework. The deep-learning-based SSM literature considered here is predominantly formulated with continuous latent states and observations. Fully discrete models, such as Hidden Markov Models, are conceptually related, but they introduce additional modeling and training considerations in the deep-learning setting and would therefore substantially broaden the scope of the review.} The index $t$ of the latent and observation processes can take values in a discrete index set $\mathcal{T}:=\{0,\cdots,T\}$, or a continuous one $\mathcal{T}:=[0,T]$.

A functional specification of the SSM relates the processes $\boldsymbol{z}_t$ and $\boldsymbol{x}_t$ through equations that depend on processes of \textit{independent shocks or noise}\footnote{Throughout, we follow the common deep learning convention of using the same notation for a random variable and its realization when the intended meaning is clear from context. We retain this convention to keep the exposition close to the literature under review.}; specifically, assuming 1-lag Markovian dynamics
\begin{subequations}
\begin{align}
\text{initial state:}\quad & \boldsymbol{z}_0 = f_0(\boldsymbol{u}_0;\theta); \label{eq35.initial-state} \\
\text{state eqn:}\quad &\boldsymbol{z}_t = f_t(\boldsymbol{z}_{t-1},\boldsymbol{u}_t;\theta), \ \ t\geq 1;  \label{eq40:state-eqn} \\
    \text{obs. eqn:}\quad &\boldsymbol{x}_t = g_t(\boldsymbol{z}_t, \bepsilon_t;\theta), \ \ t\geq 1, \label{eq50:obs-eqn}
\end{align}
\end{subequations}
where $f_0,f_t, g_t$ are deterministic multivariate functions of appropriate dimensions, and $\{\boldsymbol{u}_t\}, \{\bepsilon_t\}$ sequences of independent and identically distributed random variables of shocks/noises. $\theta$ collects the parameters associated with the state/observation equations. 
This specification is often encountered in the engineering and natural sciences literature, wherein subject matter theoretical considerations dictate the functional form of $f$ and $g$. Some illustrative examples are presented next. 
\begin{enumerate}[itemsep=0pt]
\setlength{\leftmargini}{0pt}
\item {\bf Discrete time linear Gaussian SSM.} It represents an important and extensively studied class of SSMs; see detailed expositions in the books by \cite{harvey1990forecasting,hannan2012statistical,durbin2012time}. Its specification under a time-invariant setting is given in the form of \eqref{eq35.initial-state}-\eqref{eq50:obs-eqn} with $\boldsymbol{z}_0\sim\mathcal{N}(\boldsymbol{\mu}_0,\Sigma_0)$, $f_t(\boldsymbol{z}_{t-1},\boldsymbol{u}_t)=A \boldsymbol{z}_{t-1}+\boldsymbol{u}_t$, $g_t(\boldsymbol{z}_t,\bepsilon_t)=C\boldsymbol{z}_t+ \bepsilon_t$; $A$ and $C$ are matrices of appropriate dimensions, and $\boldsymbol{u}_t \sim \mathcal{N}(0, Q), \bepsilon_t\sim \mathcal{N}(0,R)$. For this model the parameters are $\theta:=(\boldsymbol{\mu}_0,\Sigma_0, A,Q,C,R)$.
\item {\bf Nonlinear dynamical systems.} In discrete time, they are described by the set of equations \eqref{eq35.initial-state}-\eqref{eq50:obs-eqn}, while in continuous time by a set of differential equations. Concretely, a specific instance can be defined by the following coupled stochastic differential equations:
\begin{align*}
\dd\boldsymbol{z}(t) & = f(\boldsymbol{z}(t))\dd t+\sigma(\boldsymbol{z}(t))\dd \boldsymbol{w}^z(t), \\
\dd\boldsymbol{x}(t) & = g(\boldsymbol{z}(t))\dd t+\dd \boldsymbol{w}^x(t), 
\end{align*}
where $f(\cdot), \sigma(\cdot)$ and $g(\cdot)$ are measurable multivariate functions (on the appropriate filtration), while $\boldsymbol{w}^x(t)$ and $\boldsymbol{w}^z(t)$ are standard multivariate Wiener processes. 
This model serves as the theoretical foundation for the nonlinear filtering   \citep{kalman1961new,bain2009fundamentals,davis1981introduction,lototsky1997nonlinear}; see also Appendix~\ref{sec:ct-SSM}.
\end{enumerate}
The main learning task involves estimating the parameter 
$\theta$ of the model and the latent process $\boldsymbol{z}_t$, given a sequence of observations $\{\boldsymbol{x}_t, t\in\mathcal{T}\}$---collectively known as {\em system identification}. The learning of the latent process $\boldsymbol{z}_t$ itself is referred to as the {\em state estimation/identification} problem. Note that when $f$ and $g$ are known (i.e., both their functional form and the parameter $\theta$ are given)---which is often the case in engineering and other physical systems---the state identification problem has been extensively studied as an independent topic on its own in the literature.

\subsection{An overview of SSM learning} \label{sec:SSM-learning}

For an SSM learning task, one has access to observations of the system's trajectory over time. In the case of a discrete time SSM, these observations are collected directly from the recursive generation mechanism, i.e., $\boldsymbol{x}_1,\cdots,\boldsymbol{x}_T$. In the continuous case, they correspond to a realization of the underlying continuous trajectory \textit{sampled} at discrete time points; i.e., $\boldsymbol{x}_{t_1},\cdots,\boldsymbol{x}_{t_n}$, $t_i\in[0,T]$. Here and throughout, $p(\cdot)$ denotes a generic probability density or mass function. Transition, observation, and related conditional distributions are distinguished by their arguments and conditioning variables, without introducing separate notation for each. To highlight key concepts, the remainder of this section provides an overview of learning frameworks focusing on the discrete time case. The continuous time case can be handled analogously with some necessary modifications, which will be discussed in more detail as specific models are introduced in Section~\ref{sec:neural-SSM}.

The model in~\eqref{eq35.initial-state}-\eqref{eq50:obs-eqn} can be alternatively defined through the following probabilistic specification 
\begin{subequations}
\begin{align}
\text{initial state:}~~~&\boldsymbol{z}_0 \sim p_{\theta} (\boldsymbol{z}_0); \label{eq5.initial-state} \\
\text{state eqn:}~~~&\boldsymbol{z}_t \,|\,\boldsymbol{z}_{t-1}  \sim p_{\theta} (\boldsymbol{z}_t|\boldsymbol{z}_{t-1}),~t\geq 1;  \label{eq10:state-eqn} \\
\text{obs. eqn:}~~~&\boldsymbol{x}_t\,|\,\boldsymbol{z}_t \sim p_{\theta} (\boldsymbol{x}_t|\boldsymbol{z}_t),~t\geq 1.\label{eq20:obs-eqn}
\end{align}
\end{subequations}
Given an identifiable SSM\footnote{If ${\theta}\neq {\theta}'$, then $p_{{\theta}}(\mathbb{x}_{1:T}) \neq p_{{\theta}'}(\mathbb{x}_{1:T})$; namely, the set on which the two marginal likelihoods differ has a strictly positive measure.}, the static parameters $\theta$ are typically learned via maximum likelihood estimation (MLE), i.e., $\widehat{{\theta}}:=\argmax_\theta \log p_{{\theta}}(\mathbb{x}_{1:T})$. Based on the specification in \eqref{eq5.initial-state}-\eqref{eq20:obs-eqn}, the \textit{marginal likelihood} of the observed trajectory $\mathbb{x}_{1:T}$ is given by:
\begin{align}\label{eq70:likelihood-SSM}
p_{{\theta}}& (\mathbb{x}_{1:T}) = 
\int_{\mathcal{Z}^{T+1}} p_{{\theta}} (\mathbb{x}_{1:T}, \mathbb{z}_{0:T}) \dd\mathbb{z}_{0:T} = \idotsint_{\mathcal{Z}} p_{{\theta}} (\mathbb{x}_{1:T}, \mathbb{z}_{0:T}) \dd \boldsymbol{z}_0\dd\boldsymbol{z}_1\cdots\dd\boldsymbol{z}_T,
\end{align}
i.e., it is obtained by ``averaging" the joint density $p_{{\theta}}(\cdot,\cdot)$ of the observations and the latent states over all possible sequences of the latent process.

However, directly computing the likelihood in~\eqref{eq70:likelihood-SSM} poses two major challenges: (i) the high-dimensional nature of the integral over the entire latent path, whose computational complexity increases exponentially as the number of observation points $T$ increases, and (ii) the fact that for most distributions $p_{{\theta}}$, one must resort to numerical or sampling methods to evaluate the integral. Furthermore, obtaining the maximum likelihood estimate requires optimizing this complex likelihood with respect to $\theta$. Performing the likelihood evaluation in a \textit{recursive} manner addresses the first challenge by decomposing the integral, which in turn sets the stage for addressing the optimization challenge, as illustrated next.

To bypass the intractable joint integration in~\eqref{eq70:likelihood-SSM}, one exploits the conditional independence structure of the SSM to decompose the marginal likelihood sequentially. Applying the chain rule of probability yields:
\begin{align}\label{eq77:likelihood-SSM-one-step}
p_{{\theta}} (\mathbb{x}_{1:T}) &= p_{{\theta}} (\mathbb{x}_{1:T-1}) p_{{\theta}} (\boldsymbol{x}_T|\mathbb{x}_{1:T-1}) 
=\cdots=
\prod\nolimits_{t=1}^T p_{{\theta}} (\boldsymbol{x}_t | \mathbb{x}_{1:t-1}),
\end{align}
where $p_{{\theta}} (\boldsymbol{x}_t | \mathbb{x}_{1:t-1})$ is the \textit{one-step predictive likelihood} (with the convention $p_\theta(\boldsymbol{x}_1\mid \mathbb{x}_{1:0}) \coloneqq p_\theta(\boldsymbol{x}_1)$). Taking the logarithm of \eqref{eq77:likelihood-SSM-one-step} explicitly recovers the \emph{prediction-error decomposition}, $\ell(\theta) = \sum_{t=1}^T \log p_\theta(\boldsymbol{x}_t \mid \mathbb{x}_{1:t-1})$, which forms the objective function for maximum likelihood estimation.

Next, each one-step predictive likelihood can be expressed by integrating out the current latent state $\boldsymbol{z}_t$:
\begin{align} \label{eq:forward-filtering}
p_{{\theta}} (\boldsymbol{x}_t | \mathbb{x}_{1:t-1}) &= \int_{\mathcal{Z}}  p_{{\theta}} (\boldsymbol{x}_t, \boldsymbol{z}_t | \mathbb{x}_{1:t-1}) \dd \boldsymbol{z}_t 
= \int_{\mathcal{Z}} p_{{\theta}} (\boldsymbol{x}_t | \boldsymbol{z}_t)
p_{{\theta}} (\boldsymbol{z}_t | \mathbb{x}_{1:t-1}) \dd\boldsymbol{z}_t \coloneqq C_t,
\end{align}
where $p_{{\theta}}(\boldsymbol{x}_t | \boldsymbol{z}_t)$ corresponds to the probabilistic mechanism of the observation equation, and $p_{{\theta}} (\boldsymbol{z}_t | \mathbb{x}_{1:t-1})$ to the {\em posterior predictive} density of the latent state. Given the Markovian dynamics of the SSM, the posterior predictive density can be calculated using the Chapman-Kolmogorov equation; i.e., 
\begin{equation}\label{eq:predictive-c-k}
    p_{{\theta}}(\boldsymbol{z}_t | \mathbb{x}_{1:t-1})= \int_{\mathcal{Z}} p_{{\theta}}
    (\boldsymbol{z}_t | \boldsymbol{z}_{t-1}) p_{{\theta}}(\boldsymbol{z}_{t-1} | \mathbb{x}_{1:t-1}) \dd\boldsymbol{z}_{t-1}.
\end{equation}
Given the new observation $\boldsymbol{x}_t$, Bayes' rule then updates the predictive density to the \textit{filtering density}:
\begin{equation}\label{eq:filtering-equation}
    p_{\theta}(\boldsymbol{z}_t \mid \mathbb{x}_{1:t})=
    \frac{1}{C_t}
    p_{\theta}(\boldsymbol{x}_t \mid \boldsymbol{z}_t)
    p_{\theta}(\boldsymbol{z}_t \mid \mathbb{x}_{1:t-1}).
\end{equation}
The scalar $C_t$ defined in \eqref{eq:forward-filtering} acts precisely as the local normalizing constant for the filtering update. Together, the predict-update steps in \eqref{eq:predictive-c-k} and \eqref{eq:filtering-equation} form the \textit{forward filtering} recursion. By executing this recursion \textit{forward} in time, one evaluates the observed-data log-likelihood exactly via the prediction-error decomposition $\ell(\theta) = \sum_{t=1}^T \log C_t$, completely bypassing the intractable $T$-dimensional integral. 

While forward filtering leverages only past and current data (up to time $t$), obtaining optimal offline estimates of the latent states requires utilizing the full trajectory $\mathbb{x}_{1:T}$. This is achieved via the \textit{smoothing} density $p_{\theta}(\boldsymbol{z}_t \mid \mathbb{x}_{1:T})$, which is computed by running a subsequent recursion \textit{backwards} in time, outlined next.

Specifically, the \textit{smoothing problem} seeks to compute the marginal smoothing densities $p_{\theta}(\boldsymbol{z}_t \mid \mathbb{x}_{1:T})$ for all $t=1,\cdots,T$. This procedure strictly requires a completed forward filtering pass, as it relies heavily on both the filtering densities $p_{\theta}(\boldsymbol{z}_t \mid \mathbb{x}_{1:t})$ in \eqref{eq:filtering-equation} and the one-step predictive densities $p_{\theta}(\boldsymbol{z}_{t+1} \mid \mathbb{x}_{1:t})$ in \eqref{eq:predictive-c-k}.

The recursion is initialized at the final time step $T$, where the smoothing density trivially equals the final filtering density: $p_{\theta}(\boldsymbol{z}_T \mid \mathbb{x}_{1:T})$. For any time step $t < T$, we can express the marginal smoothing density by integrating the joint smoothing density of two adjacent states over the future state $\boldsymbol{z}_{t+1}$:
\begin{equation} \label{eq:marginal-smoothing-integral}
    p_{\theta}(\boldsymbol{z}_t \mid \mathbb{x}_{1:T}) = \int_{\mathcal{Z}} p_{\theta}(\boldsymbol{z}_t, \boldsymbol{z}_{t+1} \mid \mathbb{x}_{1:T}) \dd\boldsymbol{z}_{t+1}.
\end{equation}
By the chain rule of probability, the joint smoothing density factorizes as:
\begin{equation} \label{eq:joint-smoothing-factorization}
    p_{\theta}(\boldsymbol{z}_t, \boldsymbol{z}_{t+1} \mid \mathbb{x}_{1:T}) = p_{\theta}(\boldsymbol{z}_{t+1} \mid \mathbb{x}_{1:T}) \, p_{\theta}(\boldsymbol{z}_t \mid \boldsymbol{z}_{t+1}, \mathbb{x}_{1:T}).
\end{equation}
Given the SSM's Markovian structure, conditioned on the current state $\boldsymbol{z}_{t+1}$ and the past observations $\mathbb{x}_{1:t}$, the past state $\boldsymbol{z}_t$ is conditionally independent of all future observations $\mathbb{x}_{t+1:T}$. Consequently, the backward transition kernel simplifies to:
\[
    p_{\theta}(\boldsymbol{z}_t \mid \boldsymbol{z}_{t+1}, \mathbb{x}_{1:T}) = p_{\theta}(\boldsymbol{z}_t \mid \boldsymbol{z}_{t+1}, \mathbb{x}_{1:t}).
\]
Applying Bayes' rule to this reversed kernel, we can express it entirely in terms of quantities obtained during the forward pass and the system dynamics:
\begin{equation} \label{eq:backward-kernel}
    p_{\theta}(\boldsymbol{z}_t \mid \boldsymbol{z}_{t+1}, \mathbb{x}_{1:t}) = \frac{p_{\theta}(\boldsymbol{z}_{t+1} \mid \boldsymbol{z}_t) \, p_{\theta}(\boldsymbol{z}_t \mid \mathbb{x}_{1:t})}{p_{\theta}(\boldsymbol{z}_{t+1} \mid \mathbb{x}_{1:t})}.
\end{equation}
Substituting \eqref{eq:backward-kernel} back into \eqref{eq:joint-smoothing-factorization} and then into \eqref{eq:marginal-smoothing-integral} yields the fundamental \textit{backward smoothing recursion}:
\begin{equation}\label{eq:backward-smoothing-recursion}
p_{\theta}(\boldsymbol{z}_t \mid \mathbb{x}_{1:T}) = 
 p_{\theta}(\boldsymbol{z}_t \mid \mathbb{x}_{1:t}) \int_{\mathcal{Z}} \frac{p_{\theta}(\boldsymbol{z}_{t+1} \mid \boldsymbol{z}_t)}{p_{\theta}(\boldsymbol{z}_{t+1} \mid \mathbb{x}_{1:t})} p_{\theta}(\boldsymbol{z}_{t+1} \mid \mathbb{x}_{1:T}) \dd\boldsymbol{z}_{t+1}.
\end{equation} 
Notice that \eqref{eq:joint-smoothing-factorization} combined with \eqref{eq:backward-kernel} also provides the exact formula for the lag-one smoothing density:
\begin{equation*}\label{eq:lag-one-smoother}
p_{\theta}(\boldsymbol{z}_t,\boldsymbol{z}_{t+1}\mid \mathbb{x}_{1:T})
=
\frac{
p_{\theta}(\boldsymbol{z}_{t+1}\mid \boldsymbol{z}_t)\,
p_{\theta}(\boldsymbol{z}_t\mid \mathbb{x}_{1:t})\,
p_{\theta}(\boldsymbol{z}_{t+1}\mid \mathbb{x}_{1:T})
}{
p_{\theta}(\boldsymbol{z}_{t+1}\mid \mathbb{x}_{1:t})
}, 
\end{equation*}
for $t=T-1,\cdots,0$, which is particularly useful in the estimation of the $\theta$ parameter as presented in the sequel.

Hence, smoothing is a two-pass procedure: a forward pass computes the filtering
and predictive densities, and a backward pass incorporates future observations
to refine the state posterior density.

In general, smoothing is computationally more demanding than filtering. The backward pass
requires the full sequence of forward filtering densities
$p_{\theta}(\boldsymbol{z}_t\mid \mathbb{x}_{1:t})$ and predictive densities
$p_{\theta}(\boldsymbol{z}_{t+1}\mid \mathbb{x}_{1:t})$, and each step involves
integration over the latent state space. Thus, while the recursive formulation of both filtering and smoothing
avoids high-dimensional integration,
it still requires computing a sequence of lower-dimensional integrals that are rarely
available in closed form outside a few special cases--see Remark 2 in Section \ref{sec:MLE-SSM}.

These inferential quantities are precisely the building blocks for parameter
learning. On the one hand, the filtering recursion delivers the incremental
likelihood terms $C_t$, and hence the observed-data log-likelihood
$\ell(\theta)=\sum_{t=1}^T \log C_t$, which underlies the
\emph{marginalization} approach. On the other hand, the smoothing and lag-one
smoothing distributions provide the conditional expectations needed when the
latent states are treated as missing data, which leads to the
\emph{data-augmentation} approach, most notably the Expectation-Maximization (EM) algorithm. Section~\ref{sec:MLE-SSM} develops these two maximum-likelihood strategies in turn.

These same structural and computational bottlenecks in filtering and smoothing also motivate
the deep learning-based approaches discussed in Section~\ref{sec:VI-SSM}. At a conceptual level,
these methods target the same intractable posterior quantities, but replace explicit forward--backward
recursions with learned approximations. In particular, the deep learning framework leverages neural networks to flexibly parameterize the state and observation equations. Crucially, this bypasses the need to explicitly specify the underlying system dynamics mathematically---a significant advantage in many real-world applications where the true generative mechanisms are unknown or cannot be easily captured by rigid mechanistic models. To achieve this, many such approaches adopt a VAE-based learning pipeline. Rather than explicitly computing the smoothing posterior
$p_{\theta}(\mathbb{z}_{1:T}\mid \mathbb{x}_{1:T})$ through a backward recursion, an encoder network
learns an amortized variational approximation
$q_{\phi}(\mathbb{z}_{1:T}\mid \mathbb{x}_{1:T})$, while a decoder network parameterizes the generative process.
The model parameters are then learned jointly, in an end-to-end fashion, by optimizing an evidence lower bound
that serves as a tractable surrogate for the log-likelihood. In many implementations, the variational
distributions produced by the encoder and the conditional distributions specified by the decoder are taken to be Gaussian, with means and covariances (often restricted to diagonal form) parameterized by neural networks for the latent states and/or observations. This Gaussian specification simplifies both the computation of the variational evidence lower bound and the associated optimization procedure; see Section~\ref{sec:VI-SSM} for details.

\subsection{Classical approaches of maximum likelihood based learning of SSMs} 
\label{sec:MLE-SSM}

Building on the preceding discussion, maximum-likelihood estimation of the static parameters $\theta$ in an SSM is commonly pursued through two classical strategies: \emph{marginalization} and \emph{data augmentation}. In the
marginalization approach, the latent states are integrated out and one optimizes the observed-data log-likelihood directly. In the data-augmentation approach, the latent states are treated as missing variables and incorporated
explicitly into the optimization problem, through the EM algorithm.

Although these two strategies differ in how the optimization problem is formulated, both rely on the same underlying state-estimation machinery. Marginalization depends on the filtering and predictive quantities that define the observed-data likelihood, whereas data augmentation depends on the
smoothing distributions needed to form conditional expectations with respect to the latent state trajectory. We discuss these two approaches in turn.

\paragraph{Marginalization.} The first strategy estimates $\theta$ by directly maximizing the \emph{observed-data} log-likelihood. As established via the prediction-error decomposition, the objective function is exactly $\ell(\theta) = \sum_{t=1}^T \log C_t$. The maximum-likelihood estimator, $\widehat{\theta}=\argmax_{\theta}\,\ell(\theta)$, is typically obtained via an iterative numerical optimization scheme. A generic iterative update takes the form:
\[
\theta^{(k+1)} = \theta^{(k)} + \eta^{(k)}\alpha^{(k)},
\]
where $\eta^{(k)}>0$ is the step size and $\alpha^{(k)}$ is an ascent direction; e.g., the marginal gradient $\alpha^{(k)} = \nabla_\theta \ell(\theta^{(k)})$ or a Newton direction
$\alpha^{(k)} = -[\nabla_\theta^2 \ell(\theta^{(k)})]^{-1} \nabla_\theta \ell(\theta^{(k)})$. Formally,
\[
\nabla_\theta \ell(\theta)=\sum_{t=1}^T \frac{\nabla_\theta C_t}{C_t},
\]
but this expression is not especially useful computationally, because each $C_t$ depends on $\theta$ both explicitly through the observation model and implicitly through the recursively defined predictive distribution $p_\theta(\boldsymbol{z}_t\mid \mathbb{x}_{1:t-1})$. In other words, differentiating $\ell(\theta)$ directly requires differentiating through the full filtering recursion.

Under standard regularity conditions that justify differentiation under the integral sign, Fisher's identity yields the more informative representation
\begin{align*}
\nabla_\theta \ell(\theta)
&=
\int_{\mathcal{Z}^{T+1}}
\nabla_\theta \log p_\theta(\mathbb{x}_{1:T},\mathbb{z}_{0:T})\,
p_\theta(\mathbb{z}_{0:T}\mid \mathbb{x}_{1:T})
\dd\mathbb{z}_{0:T}.
\end{align*}
Using the complete-data factorization,
\begin{equation}\label{eq:comp-data-factorization}
\log p_\theta(\mathbb{x}_{1:T},\mathbb{z}_{0:T})
=
\log p_\theta(\boldsymbol{z}_0)
+\sum_{t=1}^T \log p_\theta(\boldsymbol{x}_t\mid \boldsymbol{z}_t)
+\sum_{t=1}^T \log p_\theta(\boldsymbol{z}_t\mid \boldsymbol{z}_{t-1}),
\end{equation}
the gradient decomposes at the current iterate $\theta^{(k)}$ as
{\small
\begin{equation*}
\nabla_\theta \ell(\theta) \Big|_{\theta = \theta^{(k)}}
=
\mathbb{E}_{\theta^{(k)}}\!\left[\left.\nabla_\theta \log p_\theta(\boldsymbol{z}_0)\right|_{\theta=\theta^{(k)}}\mid \mathbb{x}_{1:T}\right]
+ 
\sum_{t=1}^T
\mathbb{E}_{\theta^{(k)}}\!\left[\left.\nabla_\theta \log p_\theta(\boldsymbol{x}_t\mid \boldsymbol{z}_t)\right|_{\theta=\theta^{(k)}}\mid \mathbb{x}_{1:T}\right]
+
\sum_{t=1}^T
\mathbb{E}_{\theta^{(k)}}\!\left[\left.\nabla_\theta \log p_\theta(\boldsymbol{z}_t\mid \boldsymbol{z}_{t-1})\right|_{\theta=\theta^{(k)}}\mid \mathbb{x}_{1:T}\right],
\end{equation*}
}%
where the notation $\mathbb{E}_{\theta^{(k)}}[\,\cdot \mid \mathbb{x}_{1:T}]$ explicitly indicates that the conditional expectation is taken with respect to the smoothing posterior $p_{\theta^{(k)}}(\mathbb{z}_{0:T}\mid \mathbb{x}_{1:T})$ evaluated at the \textit{current iterate} $\theta^{(k)}$, while the gradients inside the expectation are taken with respect to $\theta$ and then evaluated at $\theta=\theta^{(k)}$.

Crucially, because the complete-data log-likelihood factorizes into terms involving at most two consecutive latent states, the high-dimensional expectations over the entire trajectory $\mathbb{z}_{0:T}$ simplify to local integrals. Consequently, exact evaluation of the marginal gradient requires only the marginal single-time smoothing densities $p_\theta(\boldsymbol{z}_t \mid \mathbb{x}_{1:T})$ (to compute the first two terms) and the lag-one joint smoothing densities $p_\theta(\boldsymbol{z}_{t-1}, \boldsymbol{z}_t \mid \mathbb{x}_{1:T})$ (to compute the transition dynamics in the final term). 

For Newton-type algorithms that exploit local curvature information, it is useful to work with the \emph{observed information matrix} $
\mathcal{I}_{\mathrm{obs}}(\theta):=-\nabla_\theta^2 \ell(\theta)$. Louis' identity \citep{Louis1982} expresses this matrix in terms of conditional moments of the complete-data derivatives:
\begin{equation*}
\mathcal{I}_{\mathrm{obs}}(\theta)
=
\mathbb{E}_\theta\!\left[
-\nabla_\theta^2 \log p_\theta(\mathbb{x}_{1:T},\mathbb{z}_{0:T})
\,\middle|\, \mathbb{x}_{1:T}
\right] -
\operatorname{Var}_\theta\!\left[
\nabla_\theta \log p_\theta(\mathbb{x}_{1:T},\mathbb{z}_{0:T})
\,\middle|\, \mathbb{x}_{1:T}
\right].
\end{equation*}
That is, the curvature of the observed-data log-likelihood can be written as the conditional expectation of the negative complete-data Hessian minus the conditional variance of the complete-data score. The second term quantifies the information loss due to the latent states being unobserved, and is therefore often referred to as a \emph{missing-information} term.

\paragraph{Data augmentation.} 
In this strategy, the latent state trajectory is treated as missing data and one works with the complete-data likelihood
\begin{equation*}
p_{{\theta}} (\mathbb{x}_{1:T}, \mathbb{z}_{0:T})=p_{{\theta}} (\boldsymbol{z}_0) \prod\nolimits_{t=1}^T p_{{\theta}} (\boldsymbol{z}_t | \boldsymbol{z}_{t-1}) \prod\nolimits_{t=1}^T p_{{\theta}} (\boldsymbol{x}_t | \boldsymbol{z}_t),
\end{equation*}
that in turn is used to estimate the model parameters ${\theta}$. The standard algorithm is EM. Given a current iterate $\theta^{(k)}$, the E-step constructs the surrogate objective
\begin{align*}
\mathcal{Q}(\theta\mid \theta^{(k)})
:=
\mathbb{E}_{\theta^{(k)}}\!\left[
\log p_\theta(\mathbb{x}_{1:T},\mathbb{z}_{0:T})
\,\middle|\, \mathbb{x}_{1:T}
\right]
=
\int_{\mathcal{Z}^{T+1}}
 \log p_\theta(\mathbb{x}_{1:T},\mathbb{z}_{0:T})\,
p_{\theta^{(k)}}(\mathbb{z}_{0:T}\mid \mathbb{x}_{1:T})
\dd\mathbb{z}_{0:T}.
\end{align*}
By the same factorization as in \eqref{eq:comp-data-factorization} we get
\begin{align*}
\mathcal{Q}(\theta\mid \theta^{(k)})
=
\mathbb{E}_{\theta^{(k)}}\!\left[\log p_\theta(\boldsymbol{z}_0)\mid \mathbb{x}_{1:T}\right] +
\sum_{t=1}^T
\mathbb{E}_{\theta^{(k)}}\!\left[\log p_\theta(\boldsymbol{x}_t\mid \boldsymbol{z}_t)\mid \mathbb{x}_{1:T}\right]
+
\sum_{t=1}^T
\mathbb{E}_{\theta^{(k)}}\!\left[\log p_\theta(\boldsymbol{z}_t\mid \boldsymbol{z}_{t-1})\mid \mathbb{x}_{1:T}\right].
\end{align*}
Thus, the E-step requires the same single-time marginal and adjacent-time joint smoothing distributions that appear in Fisher's identity. The M-step then updates the parameters by
\begin{equation*}\label{eq.150:M-step}
\theta^{(k+1)}=\arg\max_\theta \,\mathcal{Q}(\theta\mid \theta^{(k)}). 
\end{equation*}
When this maximization is available in closed form, EM is especially attractive. Otherwise, one typically uses a generalized EM step, in which $\mathcal{Q}$ is increased,  but not necessarily maximized exactly.

\begin{remark}[Operational contrast between marginalization and data augmentation]
Fisher's identity clarifies the local relationship between direct marginalization
and the missing-data strategy underlying the EM algorithm. In the EM framework,
one introduces the surrogate objective 
$\mathcal{Q}(\theta \mid \theta^{(k)})$
that is, the conditional expectation of the complete-data log-likelihood under
the current smoothing distribution. Under standard regularity conditions,
Fisher's identity implies that the observed-data score and the surrogate score
coincide at the current iterate:
\begin{equation*}
\nabla_\theta \ell(\theta)\big|_{\theta=\theta^{(k)}}
=
\nabla_\theta \mathcal{Q}(\theta\mid\theta^{(k)})\big|_{\theta=\theta^{(k)}}.
\end{equation*}
Thus, EM and direct likelihood optimization share the same first-order information at $\theta^{(k)}$.

Their update mechanisms, however, are different. In direct marginalization, one typically applies a local optimization step to the observed-data objective,
such as a gradient or Newton update based on an appropriately selected step size based on line search or trust region considerations. In EM, by contrast, the outer update is defined by maximizing the surrogate:
$\theta^{(k+1)} \in \arg\max_\theta \mathcal{Q}(\theta \mid \theta^{(k)})$.

Hence, EM replaces an explicit outer step-size choice by a surrogate maximization step. When this maximization is available in closed form, the update can be especially simple; when it is not, the M-step may still require
an inner numerical optimization routine. The distinction is therefore operational rather than informational: the two approaches are locally linked by the same first-order identity, but they package that information into different
update rules.
\end{remark}

\begin{remark}
The linear Gaussian state-space model is a canonical tractable benchmark, in which the predictive, filtering, and smoothing densities are Gaussian and admit closed-form recursive updates; see Appendix~\ref{sec:discrete-linear-Gaussian-SSM}. Other structured cases include finite-state hidden Markov models, which admit exact forward--backward recursions, and conditionally linear Gaussian models, for which Rao-Blackwellization permits partial analytical inference. Beyond such cases, exact inference is generally unavailable, particularly for nonlinear or non-Gaussian SSMs with moderate or high state or observation dimensions.

As a result, one must rely on approximate inference procedures. A first broad class consists of Gaussian approximation methods, which retain a Gaussian form for the relevant conditional distributions and propagate only their first two moments. The extended Kalman filter (EKF) \citep{anderson2005optimal}, for example, linearizes the nonlinear state-transition and observation maps about the current operating point and then applies Kalman-style recursions to the resulting local linear-Gaussian approximation. More generally, Gaussian filters approximate the predictive and filtering distributions by matching their moments, with the required integrals evaluated numerically, for instance by quadrature or Monte Carlo methods. The unscented Kalman filter (UKF) \citep{wan2000unscented,julier2004unscented}, a prominent member of this class, uses deterministically chosen ``sigma" points to approximate these moments. A brief exposition of the EKF, Gaussian filters, and the UKF is given in Appendix~\ref{sec:EKF}. Although such methods are often computationally efficient, their accuracy may deteriorate in the presence of strong nonlinearities or pronounced non-Gaussian behavior.

A second broad class consists of Sequential Monte Carlo (SMC) methods, also known as particle methods; see, for example, \cite{kitagawa1988numerical,andrieu2004particle,schon2015sequential}. In this approach, the distributions of interest are represented by a weighted sample of random particles, which are propagated over time through importance sampling and resampling steps. These methods are considerably more flexible and can be adapted to both filtering and smoothing, but their computational cost can be substantial and their performance may degrade in high-dimensional settings. Variational approaches, discussed in Section~\ref{sec:VI-SSM}, provide a further alternative by replacing the exact posterior with a tractable optimization-based approximation.
\end{remark}

\begin{remark}
In continuous time SSMs defined by ordinary or stochastic differential equations, the estimates of the latent states are obtained by solving the filter's corresponding differential equation; see, learning of latent neural ODE model and a discussion of the nonlinear filtering problem for the SDE model in Appendix~\ref{sec:ct-SSM}. This approach contrasts with the discrete time case, where prediction and filtering are distinct steps.
\end{remark}

\begin{remark}
The presentation thus far has focused on maximum likelihood-based approaches for learning the static parameters $\theta$. While Bayesian inference of the latent states (i.e., computing the filtering and smoothing posterior densities) forms the very foundation of these estimation techniques, full Bayesian estimation of the model parameters themselves, which requires specifying a prior distribution $p(\theta)$ and targeting the joint posterior $p(\theta, \mathbb{z}_{1:T} \mid \mathbb{x}_{1:T})$, is computationally particularly challenging. Consequently, it is less widely adopted in practical and high-dimensional applications, particularly for nonlinear and non-Gaussian SSMs, due to the significant computational overhead of exploring this augmented parameter-state space.

Representative works addressing full Bayesian parameter estimation for linear Gaussian SSMs include \cite{durbin2000time} and references therein, while approaches leveraging the Extended Kalman Filter to facilitate Bayesian inference in nonlinear SSMs are explored in \cite{triantafyllopoulos2021bayesian}.
\end{remark}

\subsection{Deep learning framework for SSMs}\label{sec:VI-SSM}

Deep learning (DL) offers a highly flexible framework to extend classical approaches and model more complex dynamics. Broadly, DL approaches can be categorized into two streams: (1) ``partially DL" approaches, where neural networks are used only for parameterizing selected parts of the state and/or the observation equations, yet the overall learning process follows along the lines of the classical framework (see, e.g., works in Section \ref{sec:early-DL-SSM}), and (2) ``fully DL" approaches where all model components are parameterized by neural networks, {\em and} the parameters are learned with a VAE-based pipeline. The latter has become the \textit{de facto} approach for training complex SSMs.

We begin the exposition with a brief overview of the VAE strategy \citep{kingma2013auto}, 
a general approach for training complex models with latent variables in an ``end-to-end" manner. With a slight abuse of notation, let $\boldsymbol{z}$ generically denote the latent variable and $\boldsymbol{x}$ the observed one. Their joint distribution is given by $p_{\theta}(\boldsymbol{x},\boldsymbol{z}) = p_{\theta}(\boldsymbol{x}|\boldsymbol{z})p_{\theta}(\boldsymbol{z})$, where ${\theta}$ denotes the learnable parameters of the generative process and $p_{{\theta}}(\boldsymbol{z})$ is the prior distribution, typically selected as the standard Gaussian distribution. Instead of maximizing the data log-likelihood $\log p_{{\theta}}(\boldsymbol{x})$ directly, the training objective becomes to maximize its evidence lower bound (ELBO), given by
\begin{align*}
\log p_{\theta}&(\boldsymbol{x}) \geq 
\log p_{\theta}(\boldsymbol{x}) - \sKL{q_{\phi}(\boldsymbol{z}|\boldsymbol{x})}{p_{\theta}(\boldsymbol{z}|\boldsymbol{x})} \\
&= \mathbb{E}_{q_{\phi}(\boldsymbol{z}|\boldsymbol{x})}\big( \log p_{\theta}(\boldsymbol{x}|\boldsymbol{z})\big) -\sKL{q_{\phi}(\boldsymbol{z}|\boldsymbol{x})}{p_{\theta}(\boldsymbol{z})},
\end{align*}
where $q_{\phi}(\boldsymbol{z}|\boldsymbol{x})$ is the encoder (or the recognition/inference model, equivalently), which approximates the true-but-intractable posterior distribution $p_{\theta}(\boldsymbol{z}|\boldsymbol{x})$, and $p_{\theta}(\boldsymbol{x}|\boldsymbol{z})$ is the decoder (also known as the generative model) that captures the generative process given $\boldsymbol{z}$. Both the encoder and the decoder are usually parameterized by neural networks, which are jointly trained by minimizing the negative ELBO with respect to the trainable parameters ${\phi}$ and ${\theta}$. This objective can be interpreted as the sum of the reconstruction error and the KL regularization term, with the Kullback-Leibler (KL) divergence preventing the learned encoder distribution to deviate drastically from the prior distribution.    

Next, we elaborate on how the VAE strategy is utilized for modeling sequential data. When both the observable and latent variables are {\em sequences}, denoted by $\mathbb{x}_{1:T}$ and $\mathbb{z}_{0:T}$, the ELBO for the corresponding data log-likelihood can be written as
\begin{align}\label{eqn:ELBO}
\log p_{\theta}(\mathbb{x}_{1:T})
&\geq
\mathbb{E}_{q_{\phi}(\mathbb{z}_{0:T}\mid \mathbb{x}_{1:T})}
\!\left[
\log p_{\theta}(\mathbb{x}_{1:T}\mid \mathbb{z}_{0:T})
\right] -\KL{ q_{\phi}(\mathbb{z}_{0:T}\mid \mathbb{x}_{1:T})}
{p_{\theta}(\mathbb{z}_{0:T})}.
\end{align}
Under the factorization
{\small
\begin{equation*}
q_{\phi}(\mathbb{z}_{0:T}\mid \mathbb{x}_{1:T})
= q_{\phi}(\boldsymbol{z}_0\mid \mathbb{x}_{1:T})
\prod_{t=1}^T
q_{\phi}(\boldsymbol{z}_t\mid \boldsymbol{z}_{t-1}, \mathbb{x}_{1:T}),
\end{equation*}
the ELBO admits the timestep-wise decomposition
\begin{align}\label{eqn:ELBO-timestep}
\mathbb{E}_{q_{\phi}(\mathbb{z}_{0:T}\mid \mathbb{x}_{1:T})}
\Bigg[
\sum_{t=1}^T \log p_{\theta}(\boldsymbol{x}_t\mid \boldsymbol{z}_t)
-\KL{q_{\phi}(\boldsymbol{z}_0\mid \mathbb{x}_{1:T})}
{p_{\theta}(\boldsymbol{z}_0)}
-\sum_{t=1}^T
\KL{q_{\phi}(\boldsymbol{z}_t\mid \boldsymbol{z}_{t-1},\mathbb{x}_{1:T})}
{p_{\theta}(\boldsymbol{z}_t\mid \boldsymbol{z}_{t-1})}
\Bigg].
\end{align}
}%
Here, each factor in the encoder is assumed Gaussian and serves as part of an approximation to the posterior of the latent states conditional on the observations. The decoder specifies the generative model, whose joint density factorizes as
\begin{equation*}
    p_{\theta}(\mathbb{x}_{1:T},\mathbb{z}_{0:T})= p_{\theta}(\boldsymbol{z}_0)\prod\nolimits_{t=1}^T p_{\theta}(\boldsymbol{x}_t|\boldsymbol{z}_t) p_{\theta}(\boldsymbol{z}_t|\boldsymbol{z}_{t-1}).
\end{equation*}
Specifically, under the Gaussianity assumption, the conditional distribution of an observation is $\boldsymbol{x}_{t}|\boldsymbol{z}_t\sim\mathcal{N}(\bmu^x_{t}:=\mu_{\theta}(\boldsymbol{z}_t),\Sigma_t^x:=\Sigma_{\theta}(\boldsymbol{z}_t))$, where $\mu_{\theta}(\cdot),\Sigma_{\theta}(\cdot)$ are functions of $\boldsymbol{z}_t$, and their outputs correspond to the mean and the covariance, respectively, with the latter usually assumed to be diagonal. The reconstruction error is given by the negative log-likelihood, up to an additive constant:
\begin{equation*}
    \frac{1}{2}\sum\nolimits_{t=1}^T \Big(\log \big|\Sigma_t^x\big| + \big(\boldsymbol{x}_{t} - \bmu_t^x\big)^\top \big(\Sigma_t^x\big)^{-1}\big(\boldsymbol{x}_{t} - \bmu_t^x\big)\Big);
\end{equation*}
with the corresponding ``reconstructed trajectory" given by $\widehat{\boldsymbol{x}}_{t}:=\bmu^x_{t},\forall \ t$, namely, the mean of the estimated conditional distribution. In practice, instead of obtaining the full distribution, it is often the case that only the reconstructed ``mean" is obtained, and the reconstruction error is typically calculated using the mean-squared-error between $\boldsymbol{x}_{t}$ and $\widehat{\boldsymbol{x}}_{t}$'s. Under a fixed isotropic covariance $\Sigma_t^x=\sigma^2 I_p$, minimizing the negative log-likelihood is equivalent to minimizing the mean squared error.

The encoder and decoder are trained jointly by minimizing the negative ELBO, which allows the model to simultaneously learn the latent process dynamics and the state-observation link function.

\begin{remark}
The exact implementation of the inference/recognition network that parameterizes $q_{\phi}(\boldsymbol{z}_t|\boldsymbol{z}_{t-1}, \mathbb{x})$ often varies depending on the specific task and the model architecture. In some cases, additional approximation may be introduced to simplify certain steps (e.g., see \cite{hasan2021identifying} reviewed in Section~\ref{sec:latent-neuralSDE}). Similarly, for the decoder, the conditional distribution of the observations may be operationalized differently, with the inputs to the neural networks potentially incorporating extra dependence on past observations \citep[e.g.,][]{chung2015recurrent}, albeit this deviates from the specification of observation equation.
\end{remark}

%% file: 03-neural-SSM.tex
\section{Deep Learning-based State-Space Models}\label{sec:neural-SSM}

This section reviews SSM-based modeling frameworks where deep learning techniques play a central role in facilitating the learning process. These models, similarly to the classical framework reviewed in Section~\ref{sec:intro-SSM}, incorporate a latent component that governs the system's dynamics that can be potentially complex. Observed processes are linked to the latent states via a general link function, typically parameterized by a neural network.

These deep learning-based approaches can be further segmented  into two main categories based on the latent process evolution:
(1) latent states are assumed to evolve in discrete-time \citep[e.g.,][]{krishnan2015deepkf,karl2017deep,rangapuram2018deep}; 
and (2) latent states are modeled as continuous-time processes, such as ODE-based \citep{rubanova2019latent,bilovs2021neural} and SDE-based ones \citep{jia2019neural,duncker2019learning,hasan2021identifying,deng2021continuous}. 

Throughout this section, we separate $\theta$---the parameter governing the generative model---into $\theta_f$ and $\theta_g$ that respectively represent the parameters governing the state and the observation equations; note that they also coincide with the trainable parameters in the decoder. The trainable parameters in the encoder are collectively denoted by $\phi$.


\subsection{Discrete-time neural state space models}\label{sec:discrete-deep-SSM}
\begin{algorithm*}[!htb]
\SetAlgorithmName{Exhibit}{}{}
\captionsetup{font=footnotesize}
\caption{Outline of the forward pass based on \cite{krishnan2017structured}}\label{dsspipe}
\KwIn{observed time series $\mathbb{x}:=\{\boldsymbol{x}_{1},\cdots,\boldsymbol{x}_{T}\}$; inference networks $\mu_{\phi},\Sigma_{\phi}$, state transition $f_{\theta_f},\Sigma_{\theta_f}$ and $g_{\theta_g}$ in the observation equation}\vspace*{1mm}
1. \textsc{\footnotesize~[encode]} run $\{\boldsymbol{x}_{t},t=1,\cdots,T\}$ through the inference network, which gives the approximate posterior distributions $q_\phi(\mathbb{z}|\mathbb{x})$ in the form of $\mathcal{N}(\bmu^z_{t},\Sigma^z_{t})$'s, $t=1,\cdots,T$;\vspace*{1mm}\\
2. \textsc{\footnotesize~[sampling]} sample $\widetilde{\boldsymbol{z}}_{t}$'s from the encoded distributions $\mathcal{N}(\bmu^z_{t},\Sigma^z_{t})$'s, \vspace*{1mm}\\
3. \textsc{\footnotesize~[decode]} obtain the conditional likelihood $p_{\theta}(\mathbb{x}|\mathbb{z})=\prod_{t=1}^T p(\boldsymbol{x}_t|\boldsymbol{z}_t)$ according to the observation equation, using $\widetilde{\boldsymbol{z}}_t$'s as the plug-in:
\begin{equation*}
    p(\boldsymbol{x}_{t} | \boldsymbol{z}_{t}=\widetilde{\boldsymbol{z}}_{t}) = \mathcal{N}\big(\text{mean network}_{\theta_g}(\widetilde{\boldsymbol{z}}_{t}), \text{cov network}_{\theta_g}(\widetilde{\boldsymbol{z}}_{t}) \big),~~t=1,\cdots,T.
\end{equation*}\\
4. \textsc{\footnotesize~[loss]} calculate the loss function according to~\eqref{eqn:dssELBO}. \vspace*{1mm}\\
\end{algorithm*}

These models extend classical SSMs by representing the state and the observation equations through neural network parameterization, rather than using predefined parametric forms for $f$ and $g$. Such an approach allows for flexible modeling of the complex dynamics in time series data, and the formulation can be viewed as a direct neural variant of linear SSMs.

\cite{krishnan2015deepkf,krishnan2017structured} consider a data generating process (DGP) wherein the latent state $\boldsymbol{z}_t$ is modeled as a Markov process, and its conditional distribution on its past is explicitly assumed Gaussian\footnote{The original formulation incorporates the elapsed time $\Delta t$; we suppress this dependence here to simplify the exposition and emphasize the core idea.}:
\begin{subequations}
\begin{align}
    \text{\small init. state:}~~& \boldsymbol{z}_0\sim \mathcal{N}(\boldsymbol{\mu}_0,\Sigma_0) \\
    \text{\small state eqn:}~~&\boldsymbol{z}_t\,|\boldsymbol{z}_{t-1}
    \sim \mathcal{N}\Big( f_{\theta_f}(\boldsymbol{z}_{t-1}),\Sigma_{\theta_f}(\boldsymbol{z}_{t-1})\Big)\label{eqn:dss-state} \\
    \text{\small obs. eqn:}~~&\boldsymbol{x}_t = g_{\theta_g}(\boldsymbol{z}_t) + \boldsymbol{\epsilon}_t. \label{eqn:dss-obs}
\end{align}
\end{subequations}
$f(\cdot,\cdot), \Sigma(\cdot,\cdot)$ and $g(\cdot)$ respectively define 
the dynamics of the latent process and the observation equation. To learn the parameters associated with the model, the authors consider variational inference as outlined in Section~\ref{sec:VI-SSM}, with a loss in the form of~\eqref{eqn:ELBO}. Specifically, the true posterior factorizes as $p(\mathbb{z}|\mathbb{x})=p(\boldsymbol{z}_0|\mathbb{x})\prod_{t=1}^T p(\boldsymbol{z}_t|\boldsymbol{z}_{t-1},\mathbb{x})$, where 
\begin{equation*}
    p(\boldsymbol{z}_t|\boldsymbol{z}_{t-1},\mathbb{x}):= p(\boldsymbol{z}_t|\boldsymbol{z}_{t-1},\mathbb{x}_{t:T})~\text{since}~\boldsymbol{z}_t\indep\mathbb{x}_{1:{t-1}}\,|\,\boldsymbol{z}_{t-1}.
\end{equation*}
The encoder distribution $q_{\phi}$ is selected to approximate this true posterior, namely,
\begin{equation}\label{eqn:dss-approx-pos}
q_{\phi}(\mathbb{z}|\mathbb{x}) := q_{\phi}(\boldsymbol{z}_0|\mathbb{x})\prod\nolimits_{t=1}^T q_{\phi}(\boldsymbol{z}_t|\boldsymbol{z}_{t-1},\mathbb{x}_{t:T}),
\end{equation}
where
\begin{equation}\label{eqn:dss-inference-net}
    q_{\phi}(\boldsymbol{z}_t|\boldsymbol{z}_{t-1},\mathbb{x}) = \mathcal{N}\Big(\mu_{\phi}(\boldsymbol{z}_{t-1},\mathbb{x}_{t:T}), \Sigma_{\phi}(\boldsymbol{z}_{t-1},\mathbb{x}_{t:T})\Big).
\end{equation}
Both $\mu_{\phi}(\cdot)$ and $\Sigma_{\phi}(\cdot)$ are parametrized by neural networks with the output of $\Sigma_{\phi}(\cdot)$ further assumed diagonal.  The ELBO loss can be written as
{\small
\begin{align}\label{eqn:dssELBO}
\sum\nolimits_{t=1}^T \mathbb{E}_{q_{\phi}(\boldsymbol{z}_{t}|\mathbb{x})} \log p_{\theta_g}(\boldsymbol{x}_t|\boldsymbol{z}_t) - \KL{q_{\phi}(\boldsymbol{z}_0|\mathbb{x})}{p(\boldsymbol{z}_0)}
-\sum\nolimits_{t=1}^T \mathbb{E}_{q_{\phi}(\boldsymbol{z}_{t-1}|\mathbb{x})}\Big[\KL{q_{\phi}(\boldsymbol{z}_{t}|\boldsymbol{z}_{t-1},\mathbb{x}_{t:T})}{p_{\theta_f}(\boldsymbol{z}_t|\boldsymbol{z}_{t-1})}\Big],
\end{align}
}%
where $p_{\theta_f}(\boldsymbol{z}_t|\boldsymbol{z}_{t-1})$, $p_{\theta_g}(\boldsymbol{x}_t|\boldsymbol{z}_t)$ are decoder distributions dictated by the latent state dynamics~\eqref{eqn:dss-state} and the observation equation specification~\eqref{eqn:dss-obs}, and the functions involved are parameterized by neural networks. Exhibit~\ref{dsspipe} outlines the forward pass of the training pipeline.

\begin{remark} The approximate posterior distribution in~\eqref{eqn:dss-approx-pos}---operationalized via the inference network in~\eqref{eqn:dss-inference-net}--corresponds to a variant of the smoothing step in the classical Kalman filter; see, e.g., \eqref{eq:marginal-smoothing-integral} and references in \cite{briers2010smoothing}. In practice, other choices of the inference network have been considered, such as the ones based on the entire observed trajectory $\mathbb{x}$ that correspond to a fixed interval smoother, namely, $\mu_{\phi}$ and $\Sigma_{\phi}$ are in the form of $\mu_{\phi}(\boldsymbol{z}_{t-1}, \mathbb{x})$ and $\Sigma_{\phi}(\boldsymbol{z}_{t-1},\mathbb{x})$, or those depending on left-only information (i.e., $\mu_{\phi}(\boldsymbol{z}_{t-1}, \mathbb{x}_{1:t})$ and $\Sigma_{\phi}(\boldsymbol{z}_{t-1},\mathbb{x}_{1:t})$). Other mean-field type specifications let the inference network only operate on (a subset of) $\mathbb{x}$ while ignoring $\boldsymbol{z}_{t-1}$, that is, $\mu_{\phi}(\mathbb{x}),\Sigma_{\phi}(\mathbb{x})$ \citep{krishnan2017structured}, or the ``neighborhood-only" ones with $\mu_{\phi}(\boldsymbol{x}_t)$ and $\Sigma_{\phi}(\boldsymbol{x}_{t},\boldsymbol{x}_{t-1})$ \citep{gao2016linear, archer2016black}. 
\end{remark}

In the method reviewed above, the state transition enters into the loss function only through the KL term as $p_{\theta_f}(\boldsymbol{z}_t|\boldsymbol{z}_{t-1})$. In the decoding step, the conditional likelihood $p_{\theta_g}(\boldsymbol{x}_t|\boldsymbol{z}_t)$ is computed by plugging in the corresponding $\widetilde{\boldsymbol{z}}_t$'s, effectively sidelining the role of the state transition. To address this, \cite{karl2017deep} propose embedding the state transition into the decoding step itself. Concretely, instead of directly using sampled $\widetilde{\boldsymbol{z}}_{t-1}$'s to obtain the conditional likelihood, the state transition equation generates $\widehat{\boldsymbol{z}}_{t}$'s, which are subsequently used to compute  the conditional likelihood. This approach preserves the generative model's full dynamics during decoding, enabling the gradient from the reconstruction error term to backpropagate through the state transition. It is operationalized leveraging a similar idea to that of the reparametrization trick in VAEs \citep{kingma2013auto}: a ``stochastic" parameter $\boldsymbol{\beta}_t$ is introduced, which encompasses both the unknown parameters in the transition equation and the shock (namely, $\boldsymbol{u}_t$). The inference network, in this case, encodes the posterior distribution of $\boldsymbol{\beta}_t$, rather than that of $\boldsymbol{z}_t$. During the sampling stage, stochastic parameters are drawn from the inference network-learned encoder distributions, which are then passed to the decoding stage and enable deterministic state transition yielding the $\widehat{\boldsymbol{z}}_t$'s. Empirically, enforcing state transitions within the generative process enables this method to capture latent system dynamics fairly accurately, although specific parameterization choices are often application-dependent.

Several other concurrent works, such as \cite{fraccaro2017disentangled, yingzhen2018disentangled} that are reviewed in \cite{MAL-089}, explore variations of discrete-time ``deep" state space models tailored to specific tasks. Note that 
the main focus of \cite{MAL-089} is to consider VAE as a learning paradigm for dynamical systems that evolve at discrete time points; hence it also includes selected RNN-style ``hybrid" models, featuring a \textit{deterministic} internal state sequence $\{\boldsymbol{h}_t\}$ that propagates forward the system information and interacts with the \textit{stochastic} state $\boldsymbol{z}_t$. For further details, interested readers are directed to references therein.


\subsection{Latent neural ODEs}\label{sec:neuralODE}

Before introducing the latent neural ODE framework, we first present the neural ODE one as a general tool, in which the hidden unit dynamics of a neural network are modeled with ordinary differential equations (ODE). As it will be seen in the sequel, the technical components adopted in neural ODE are extensively used for latent neural ODE-based modeling.

\paragraph{Preliminaries: neural ODEs.} The neural ODE model, introduced by \cite{chen2018neural}, models the instantaneous change of the hidden units in a neural network using a function $f_\theta(\mathbf{h}_t, t)$. The motivation comes from the fact that neural networks can be viewed as a sequence of transformations applied to the hidden units, based on which connections can be established to the discretization of continuous transformations.

To set the stage, consider a neural network-based supervised learning setting where the output $\boldsymbol{y}\approx \text{DNN}(\boldsymbol{x})$, with $\text{DNN}(\cdot)$ denoting a deep neural network. With a slight abuse of notation, let $t$ index the layers of the network. For a neural network with layers $\{0,...,T\}$, let $\mathbf{h}_t$ be the vector of hidden neurons at layer $t$; a forward pass of the neural network effectively conducts the transformation $\boldsymbol{x}\rightarrow \mathbf{h}_0 \rightarrow \cdots \rightarrow \mathbf{h}_T \rightarrow \boldsymbol{y}$. In particular, for a residual network-type architecture \citep[ResNet;][]{he2016deep} wherein all hidden layers have the same size, the sequential update between hidden layers is given by $\mathbf{h}_{t+1} = \mathbf{h}_{t} + f_\theta(\mathbf{h}_{t},t)$.
This recursive update is similar to the Euler discretization scheme used to numerically obtain solutions of ODEs.

As the step size approaches zero, the dynamics of the hidden units---now viewed as a continuous process---can be represented through the following ODE: $\frac{\dd \mathbf{h}(t)}{\dd t} = f_{\theta}(\mathbf{h}(t),t)$, with the dependency of $f_{\theta}(\cdot,\cdot)$ on $t$ made explicit. The quantity of interest is given by $\mathbf{h}_{t_1}$ for some $t_1$, which can be interpreted as the ``final" hidden unit, which is subsequently mapped to the output $\boldsymbol{y}$ of ultimate interest; this mapping step is usually through a small shallow network. In particular, $\mathbf{h}_{t_1}$ can be expressed as $\mathbf{h}_{t_1}:=\mathbf{h}_{t_0}+\int_{t_0}^{t_1} f_\theta(\mathbf{h}(t),t)\dd t$, with initial condition $\mathbf{h}(t_0)=\mathbf{h}_{t_0}$ that corresponds to some initial transformation of the input $\boldsymbol{x}$\footnote{In the common supervised learning context, let $(\boldsymbol{x},\boldsymbol{y})$ denote the input-output pair. As a concrete example in the context of image classification, $\mathbf{h}_{t_0}= \mathbf{h}(t_0)$ can be the output of the initial convolution layer, $\mathbf{h}_{t_1}$ is the output after a series of such operations, and is passed to the final linear classifier (e.g., linear and softmax) after average pooling.}. The integration is typically performed using an off-the-shelf ODE solver, which is treated as a black box. In practice, the state at $t_1$ is computed as $\mathbf{h}_{t_1}=\text{ODESolve}\big(\mathbf{h}_{t_0}, f_\theta, (t_0, t_1)\big)$. The end-to-end training involves optimizing the loss with respect to the parameters  $\theta,t_0,t_1$ (and potentially the parameters of the final small shallow network that maps $\mathbf{h}_{t_1}$ to $\boldsymbol{y}$). Given the continuous nature of the hidden units in this framework, backpropagation is carried out using the adjoint method \citep{pontryagin1962math}, as detailed in \cite[Appendix B]{chen2018neural}.

\begin{algorithm*}[!hbt]
\SetAlgorithmName{Exhibit}{}{}
\captionsetup{font=footnotesize}
\caption{Outline of the forward pass based on latent neural ODE \citep{rubanova2019latent}}\label{odepipe}
\KwIn{observed time series $\mathbb{x}:=\{\boldsymbol{x}_{t_0},\cdots,\boldsymbol{x}_{t_n}\}$ }\vspace*{1mm}
1. \textsc{\footnotesize~[encode]} run $\{(\boldsymbol{x}_{t_i}, t_i),i=0,\cdots,n\}$ through the inference network backwards in time, which gives rise to the mean and variance of the approximate posterior $q_\phi(\boldsymbol{z}_{t_0}|\mathbb{x},\{t_0,\cdots,t_n\})$ in the form of $\mathcal{N}(\bmu^z_{t_0},\Sigma^z_{t_0})$;\vspace*{1mm}\\
2. \textsc{\footnotesize~[sampling]} sample $\widetilde{\boldsymbol{z}}_{t_0}$ from the encoded distribution $\mathcal{N}(\bmu^z_{t_0},\Sigma^z_{t_0})$; \vspace*{1mm}\\
3. \textsc{\footnotesize~[decode]}\\
\quad (3.1) solve the initial value problem with $\widehat{\boldsymbol{z}}_{t_0}= \widetilde{\boldsymbol{z}}_{t_0}$ and obtain $\widehat{\boldsymbol{z}}_{t_1},\cdots,\widehat{\boldsymbol{z}}_{t_n}$ based on the ODE:
\begin{equation*}
    \widehat{\boldsymbol{z}}_{t_1},\cdots,\widehat{\boldsymbol{z}}_{t_n} = \text{ODESolve}\big(\widetilde{\boldsymbol{z}}_{t_0},f_{\theta_f}, (t_0,t_1,\cdots,t_n)\big);
\end{equation*}\\
\quad (3.2) obtain the conditional data likelihood based on the observation equation in~\eqref{eqn:obs-ODE}:
\begin{equation*}
    p(\boldsymbol{x}_{t_i} | \boldsymbol{z}_{t_i}=\widehat{\boldsymbol{z}}_{t_i}) = \mathcal{N}\big(\text{mean network}_{\theta_g}(\widehat{\boldsymbol{z}}_{t_i}), \text{cov network}_{\theta_g}(\widehat{\boldsymbol{z}}_{t_i}) \big),~~i=0,\cdots,n.
\end{equation*}\\
4. \textsc{\footnotesize~[loss]} $\sum_{i=0}^n\mathbb{E}_{q_\phi(\boldsymbol{z}_{t_0}|\mathbb{x},\{t_0,\cdots,t_n\})}\log p_\theta(\boldsymbol{x}_{t_i}|\boldsymbol{z}_{t_i})-\sKL{q_\phi(\boldsymbol{z}_{t_0}|\mathbb{x},\{t_0,\cdots,t_n\})}{p(\boldsymbol{z}_{t_0})}$.\vspace*{1mm}\\
\end{algorithm*}

The general setup of neural ODEs builds upon the connection between residual networks (ResNet) and the Euler discretization, treating the hidden units/layers as a continuous process. This approach can be viewed as a continuous approximation of traditional neural network architectures, where instead of discrete layers indexed by integers, the propagation of information between hidden units occurs in a continuous manner. While this framework can potentially reduce memory requirements and enhance computational efficiency, it may seem unintuitive from a purely modeling perspective. However, in time series settings where there is an actual underlying process evolving over time, the neural ODE framework becomes particularly relevant especially when combined with a posited latent process, as mentioned briefly in \cite[Section 5]{chen2018neural} and investigated in-depth in \cite{rubanova2019latent}.  

\paragraph{Back to the latent neural ODE model.} Let $t_0,\cdots,t_n$ denote the observation timestamps. Under the latent neural ODE framework, the following DGP is assumed for the system:  
\begin{align}
\text{initial state:}\quad &\boldsymbol{z}(t_0) \sim p(\boldsymbol{z}_{t_0}), \nonumber\\
\text{state eqn:}\quad& \dd \boldsymbol{z}(t)/\dd t = f_{\theta_f}(\boldsymbol{z}(t)),\nonumber \\
\text{obs. eqn:}\quad & \boldsymbol{x}(t) = g_{\theta_g}(\boldsymbol{z}(t)) + \boldsymbol{\epsilon}(t) \label{eqn:obs-ODE}.
\end{align}
The latent process $\boldsymbol{z}(t)$ is modeled according to an ODE governed by a time invariant function $f_{\theta_f}(\cdot)$\footnote{While more general neural ODE formulations may allow explicit time dependence in the vector field, we follow the autonomous formulation adopted in the original Latent ODE model of \cite{rubanova2019latent}}, which is parameterized through some neural network with parameters $\theta_f$. The observation model defined by the link function $g_{\theta_g}(\cdot)$ is parameterized through some shallow neural network or even a simple linear layer. The forward pass of the VAE-based training pipeline is summarized in Exhibit~\ref{odepipe}, and the prior distribution $p(\boldsymbol{z}_{t_0})$ is typically selected to be a multivariate standard Gaussian. 

Referring to the generic description in~\eqref{eqn:ELBO}, it can be easily seen that in the current setting, only the KL term associated with the initial state $\boldsymbol{z}_0$ is present, due to the noiseless nature of the assumed state dynamics. Specifically, the encoder compresses the observed trajectory $\mathbb{x}$ into $\boldsymbol{z}_{t_0}$, and the ODE solver solves the corresponding initial-value problem, generating the $\boldsymbol{z}_{t_i}$'s during decoding. The authors further discuss different choices for the encoder module (i.e., the recognition network), including using an RNN or an ODE-RNN; for additional details, we refer interested readers to \cite{rubanova2019latent}. 

Within the same modeling framework wherein the latent dynamics (state equation) are specified through an ODE, subsequent work by \cite{bilovs2021neural} considers directly modeling the \textit{solution} of the ODE through a neural network (referred to as the ``neural flows"). Specifically, let $F_{\theta_f}(t,\boldsymbol{z}_{t_0})$ be the solution to the initial value problem 
\begin{equation*}
    \frac{\dd \boldsymbol{z}(t)}{\dd t} = f(\boldsymbol{z}(t),t), \quad \boldsymbol{z}(t_0) = \boldsymbol{z}_{t_0}; 
\end{equation*}
the solution is parameterized by a neural network, under the assumption  that it is smooth and invertible. These conditions ensure the solution's uniqueness. In this approach, the neural network directly learns the function $F_{\theta_f}(\cdot,\cdot)$ which is the solution of an (unknown) ODE that captures the dynamics, rather than the instantaneous changes $f$. The training pipeline is similar to the one outlined above; however, at the decoding stage, instead of using an ODE solver to obtain the $\boldsymbol{z}_{t_i}$'s, they are directly computed as $\boldsymbol{z}_{t_i}:=F_{\theta_f}(t_i,\boldsymbol{z}_{t_0})$, effectively bypassing the need for a numerical ODE solver. Empirically, the neural flow-based approach achieves comparable performance to the latent neural ODE on several real datasets in terms of model accuracy, but with significantly reduced wall-clock time during training.


\subsection{Latent neural SDEs}
\label{sec:latent-neuralSDE}

In this subsection, we present the formulation of neural SDEs that incorporate stochastic dynamics, along with their model training pipeline. Further, we highlight key differences in encoding and decoding procedures between neural ODEs and SDEs.

The extension of the neural ODE framework to an SDE setting has been considered in a series of concurrent papers across diverse contexts. For example, \cite{tzen2019neural} shows that similar to neural ODEs, neural SDEs can be viewed as the limiting regime of Deep Latent Gaussian Models \citep{rezende2014stochastic} as the number of layers goes to infinity, with the step size and layer-to-layer transformation noise variance going to zero \citep[see also][]{peluchetti2020infinitely}. \cite{Liu_2020_CVPR} considers SDE as a ``noisy" counterpart to ODE, which coincides with certain regularization mechanisms and sometimes offers greater robustness than a deterministic formulation; see also \cite{wang2019resnets}. \cite{li2020scalable} generalizes the adjoint sensitivity method \citep[e.g.,][]{pearlmutter1995gradient} to enable backpropagation through an SDE solver for neural network training. In these works, instead of using a deterministic function to model the continuous process, a diffusion term that carries stochasticity is additionally incorporated.

Representative work on using neural SDEs to model latent processes in the context of time series include \cite{duncker2019learning,jia2019neural,hasan2021identifying,deng2021continuous}. We anchor our review based on \cite{hasan2021identifying} which considers a more ``standard" setup, and discuss the other approaches in Section~\ref{sec:neuralSDE-others}. 

To this end, under the latent neural SDE framework, the system dynamics are defined as follows:
\begin{subequations}
\begin{align}
\text{state eqn:}~&\dd \boldsymbol{z}(t) = f_{\theta_f}\big(\boldsymbol{z}(t)\big) \dd t + \sigma\big(\boldsymbol{z}(t)\big)\dd \boldsymbol{w}(t), \label{eqn:latentSDE}\\
\text{obs. eqn:}~&\boldsymbol{x}(t) = g_{\theta_g}(\boldsymbol{z}(t)) + \bepsilon(t), \label{eqn:obs-SDE}
\end{align}
\end{subequations}
where $\boldsymbol{w}(t)$ denotes a $d$-dimensional standard Wiener process; $f:\mathbb{R}^d\mapsto \mathbb{R}^d$ is a time-invariant drift function, $\sigma:\mathbb{R}^{d}\mapsto \mathbb{R}^{d\times d}$ the diffusion coefficient and $\boldsymbol{\epsilon}(t)$ an i.i.d. noise process. The $f$ and $\sigma$ functions are assumed to be globally Lipschitz and satisfy a linear growth condition ensuring that the state equation admits a unique solution (see discussion in Appendix~\ref{sec:ct-SSM}). The system is identifiable only up to an equivalence
class of latent representations related through invertible transformations. Specifically, for a sufficiently smooth invertible mapping $\varphi$, the transformed process $\widetilde{\boldsymbol{z}}(t)=\varphi(\boldsymbol{z}(t))$
satisfies an SDE whose drift and diffusion coefficients $(\widetilde{f},\widetilde{\sigma})$ are determined by It\^o's formula, while the observation mapping becomes $\widetilde{g}=g\circ\varphi^{-1}$.
Furthermore, under the additional conditions specified in \citet[Theorem 2]{hasan2021identifying},
there exists an equivalent representation with identity diffusion, i.e., $\widetilde{\sigma}(\boldsymbol{z})=I_d,\forall \ \boldsymbol{z}\in\mathbb{R}^d$. Under these conditions, one may therefore focus on
latent dynamics governed by a ``simplified" SDE with identity diffusion:
$\dd \boldsymbol{z}(t) = f_{\theta_f}\big(\boldsymbol{z}(t)\big)\dd t + \dd \boldsymbol{w}(t)$, and on recovering $(f,g)$ with learnable parameters $\theta:=\{\theta_f,\theta_g\}$.

Next, we briefly describe the construction of the encoder, the decoder and the corresponding training pipeline. First, note that for any pairs of $(\boldsymbol{x}_{t},\boldsymbol{x}_{t+\Delta t})$ and $(\boldsymbol{z}_{t},\boldsymbol{z}_{t+\Delta t})$, 
\begin{align}\label{eqn:sde-posterior}
p\big( \boldsymbol{z}_{t},&\boldsymbol{z}_{t+\Delta t} \,|\, \boldsymbol{x}_{t},\boldsymbol{x}_{t+\Delta t}\big) = p\big(\boldsymbol{z}_{t+\Delta t}|\boldsymbol{x}_{t+\Delta t}, \boldsymbol{z}_{t}\big)p\big(\boldsymbol{z}_t|\boldsymbol{x}_t, \boldsymbol{x}_{t+\Delta t}\big).
\end{align}
Leveraging \eqref{eqn:sde-posterior}, the {\em true} posterior distribution can be {\em approximately} decomposed as\footnote{We elaborate on how this is derived. First, note that the 1st term on the RHS of~\eqref{eqn:sde-posterior} uses the fact that $\boldsymbol{z}_{t+\Delta t}\indep \boldsymbol{x}_{t}\,|\,\boldsymbol{z}_t$. For the 2nd term on the RHS, \cite[equation 11]{hasan2021identifying} in fact uses an approximation $p\big(\boldsymbol{z}_t|\boldsymbol{x}_t,\boldsymbol{x}_{t+\Delta t}\big)\approx p\big(\boldsymbol{z}_t|\boldsymbol{x}_t\big)$ which disregards the dependency on $\boldsymbol{x}_{t+\Delta t}$ and gives $p\big( \boldsymbol{z}_{t},\boldsymbol{z}_{t+\Delta t} \,|\, \boldsymbol{x}_{t},\boldsymbol{x}_{t+\Delta t}\big) \approx p\big(\boldsymbol{z}_{t+\Delta t}|\boldsymbol{x}_{t+\Delta t}, \boldsymbol{z}_{t}\big)p\big(\boldsymbol{z}_t|\boldsymbol{x}_t\big)$. Applying this $(t,\Delta t)$ ``paired relationship" on consecutive observations gives $p_\theta(\mathbb{z}|\mathbb{x})
=p_{\theta}\big(\boldsymbol{z}_{t_0}\,|\,\mathbb{x}\big)\prod\nolimits_{i=1}^n p_{\theta}\big(\boldsymbol{z}_{t_i}\,|\,\{\boldsymbol{z}_{t_j}\}_{j=0}^{i-1},\mathbb{x}\big)
=p_{\theta}\big(\boldsymbol{z}_{t_0}\,|\,\mathbb{x}\big)\prod\nolimits_{i=1}^n p_{\theta}\big(\boldsymbol{z}_{t_i}\,|\,\boldsymbol{z}_{t_{i-1}},\{\boldsymbol{x}_{t_j}\}_{j=i}^{n}\big)
\approx p_{\theta}\big(\boldsymbol{z}_{t_0}\,|\,\boldsymbol{x}_{t_0}\big)\prod\nolimits_{i=1}^n p_{\theta}\big(\boldsymbol{z}_{t_i}\,|\,\boldsymbol{z}_{t_{i-1}},\boldsymbol{x}_{t_i}\big).$
\label{foot}}
\begin{align}\label{eqn:approx-SDE-encoder-1}
p_\theta(\mathbb{z}|\mathbb{x}) 
\approx p_\theta(\boldsymbol{z}_{t_n}|\boldsymbol{z}_{t_{n-1}},\boldsymbol{x}_{t_n})\cdots p_\theta(\boldsymbol{z}_{t_1}|\boldsymbol{z}_{t_0},\boldsymbol{x}_{t_1})p(\boldsymbol{z}_{t_0}|\boldsymbol{x}_{t_0}),
\end{align}
where $\boldsymbol{z}_{t_i}$ depends only on $(\boldsymbol{z}_{t_{i-1}},\boldsymbol{x}_{t_{i}}),\forall \ i$. The authors consider a further approximation by using an encoder $q_\phi(\boldsymbol{z}_{t}|\boldsymbol{x}_t)$ that omits the ``autoregressive" component, allowing $\boldsymbol{z}_t$ to be inferred solely based on $\boldsymbol{x}_t$. This approximation is reasonably tight as long as the noise term $\boldsymbol{\epsilon}_t$ in the observation equation is small, and it becomes exact if $\boldsymbol{\epsilon}_t=0$ and $g(\cdot)$ is injective. Such an approximation gives rise to the following encoder in practice:
\begin{align}
&q_\phi(\mathbb{z}|\mathbb{x}) \approx \prod\nolimits_{i=0}^n q_\phi(\boldsymbol{z}_{t_i}|\boldsymbol{x}_{t_i}); \label{eqn:approx-SDE-encoder-2}\\
&\boldsymbol{z}_{t_i}|\,\boldsymbol{x}_{t_i} \sim \mathcal{N}\Big(\bmu^z_{t_i}:=\mu_{\phi}(\boldsymbol{x}_{t_i}), \Sigma^z_{t_i}:=\Sigma_{\phi}(\boldsymbol{x}_{t_i}) \Big),\nonumber
\end{align}
where $\mu_{\phi}(\cdot),\Sigma_\phi(\cdot)$ are parameterized by neural networks. On the other hand, the joint distribution factorizes as
\begin{align*}
p\big((\boldsymbol{x}_{t},&\boldsymbol{x}_{t+\Delta t}), (\boldsymbol{z}_{t},\boldsymbol{z}_{t+\Delta t})\big) = p\big(\boldsymbol{x}_{t+\Delta t}|\boldsymbol{z}_{t+\Delta t}\big)p\big(\boldsymbol{x}_{t}|\boldsymbol{z}_t\big)p\big(\boldsymbol{z}_{t+\Delta t}|\boldsymbol{z}_t\big) p\big(\boldsymbol{z}_t\big);  
\end{align*}
consequently, the joint generative distribution admits the following factorization
\begin{align*}
    p_\theta(\mathbb{x}, \mathbb{z})
    &= p(\boldsymbol{z}_{t_0})
    \prod\nolimits_{i=0}^{n}
    p_{\theta_g}(\boldsymbol{x}_{t_i} \mid \boldsymbol{z}_{t_i})
    \prod\nolimits_{i=1}^{n}
    p_{\theta_f}(\boldsymbol{z}_{t_i} \mid \boldsymbol{z}_{t_{i-1}}),
\end{align*}
where $p_{\theta_f}(\boldsymbol{z}_{t_i}|\boldsymbol{z}_{t_{i-1}})$ describes the SDE dynamics, and $p_{\theta_g}(\boldsymbol{x}_{t_i}|\boldsymbol{z}_{t_i})$ the observation model. Specifically, if the $t_i$'s are equally spaced with $\Delta t := t_i - t_{i-1}$ for all $i$, the Euler--Maruyama approximation gives the Gaussian transition
\begin{equation}\label{eqn:znext}
\boldsymbol{z}_{t+\Delta t}|\,\boldsymbol{z}_t \sim  \mathcal{N}\Big(\boldsymbol{z}_t + f_{\theta_f}(\boldsymbol{z}_t)\Delta t, (\Delta t) I_d\Big),
\end{equation}
where the discrete update uses $\boldsymbol{z}_{t+\Delta t}\approx\boldsymbol{z}_t + f_{\theta_f}(\boldsymbol{z}_t)\Delta t + \boldsymbol{w}_{t+\Delta t} - \boldsymbol{w}_t$, and $(\boldsymbol{w}_{t+\Delta t} - \boldsymbol{w}_t)$ is distributed as a multivariate Gaussian with variance $(\Delta t) I_d$.
The forward pass is summarized in Exhibit~\ref{sdepipe}. 
\begin{algorithm*}[!hbt]
\SetAlgorithmName{Exhibit}{}{}
\captionsetup{font=footnotesize}
\small
\caption{Outline of the approximate pairwise VAE training for the latent neural SDE} \citep{hasan2021identifying}\label{sdepipe}
\KwIn{observed time series $\boldsymbol{x}_{t_0},\cdots,\boldsymbol{x}_{t_n}$ }\vspace*{1mm}
1. \textsc{\footnotesize~[encode]} run $\{\boldsymbol{x}_{t_i},i=0,\cdots,n\}$ through the inference network in parallel w.r.t. $i$, which gives rise to the mean and variance $(\bmu^z_{t_i},\Sigma^z_{t_i})$'s of the approximate posteriors $q_{\phi}(\boldsymbol{z}_{t_i}|\boldsymbol{x}_{t_i})$'s;\vspace*{1mm}\\
2. \textsc{\footnotesize~[sampling]} sample $\widetilde{\boldsymbol{z}}_{t_i}$'s from the corresponding encoded distributions $\mathcal{N}(\bmu^z_{t_i},\Sigma^z_{t_i})$; \vspace*{1mm}\\
3. \textsc{\footnotesize~[decode]}\\
\quad (3.1) for each sampled $\widetilde{\boldsymbol{z}}_{t_i},i=0,\cdots,(n-1)$, obtain the ``next" $\widehat{\boldsymbol{z}}_{t_{i+1}}$ based on~\eqref{eqn:znext} via a stochastic draw;\\
\quad (3.2) obtain the conditional observation distribution $p(\boldsymbol{x}_{t_i}|\boldsymbol{z}_{t_i}),i=1,\cdots,n$ based on the observation equation in~\eqref{eqn:obs-SDE}:
\begin{equation*}
    p(\boldsymbol{x}_{t_i} | \boldsymbol{z}_{t_i}=\widehat{\boldsymbol{z}}_{t_i}) = \mathcal{N}\big(\text{mean network}_{\theta_g}(\widehat{\boldsymbol{z}}_{t_{i}}), \text{cov network}_{\theta_g}(\widehat{\boldsymbol{z}}_{t_{i}}) \big),~~i=1,\cdots,n.
\end{equation*}\\
4. \textsc{\footnotesize~[loss]} The training objective is based on a pairwise ELBO
for consecutive observations, incorporating the factorized recognition model~\eqref{eqn:approx-SDE-encoder-2},
the observation likelihood, and the SDE transition density~\eqref{eqn:znext}. Its practical evaluation
involves additional approximations; see \citet[Appendix A]{hasan2021identifying} for details.\vspace*{1mm}\\
\end{algorithm*}

\subsubsection{Other latent neural SDE-based models}\label{sec:neuralSDE-others}

Next, we briefly discuss additional representative works that model latent dynamics with SDEs. 

In \cite{duncker2019learning}, the observable $\boldsymbol{x}_t$'s depend on the latent states through a generalized linear link function with a known parametric form (but unknown parameters), and the latent dynamics follow an SDE, wherein the drift term is modeled by a Gaussian process. 

In \cite{jia2019neural}, a hybrid flow-plus-jump system is proposed to handle events with either discrete types or continuous-value features. In that setting, the latent state $\boldsymbol{z}(t)$ governs the system, while $\boldsymbol{x}_t$ corresponds to the ``embedding" vector of the observed event sampled from $\boldsymbol{x}(t)\sim p(\boldsymbol{x}|\boldsymbol{z}(t))$\footnote{In the case of discrete event types, the support of $\boldsymbol{x}(t)$ is a finite set of one-hot encodings; in the case where the events are characterized by real-valued features, $p(\boldsymbol{x}|\boldsymbol{z}(t))$ is parameterized with a mixture of Gaussian whose parameters depend on $\boldsymbol{z}(t)$.}. The probability of an event happening satisfies
\begin{align*}
& \mathbb{P}\big(\text{event happens in}~[t,t+\dd t)\,|\,\mathcal{H}_t\big) := \lambda(\boldsymbol{z}(t))\dd t,\qquad\text{where}~~\mathcal{H}:=\{(t_i,\boldsymbol{x}_{t_i})\};
\end{align*}
$\lambda(\boldsymbol{z}(t))$ is the conditional intensity function and $\mathcal{H}_t$ the subset of events up to $t$ (excluded). Further, let $N(t)$ be the number of events up to time $t$, and the latent dynamics is given as follows, with a flow component $f_\theta(\cdot,\cdot)$ and a jump one $\nu_\theta(\cdot,\cdot,\cdot)$ (for a technical presentation of SDEs with jumps see \citep{bass2004stochastic}):
\begin{equation}\label{eqn:zjump}
    \dd \boldsymbol{z}(t) = f_\theta\big(\boldsymbol{z}(t),t\big)\dd t + \nu_{\theta}\big(\boldsymbol{z}(t),\boldsymbol{x}(t),t\big)\dd N(t).
\end{equation}
Note that with the dynamics postulated in~\eqref{eqn:zjump}, the latent process ``interacts" with the observable process through the jump component, evolving {\em deterministically} until a stochastic event occurs. Notably, the learning of various system components (e.g., $\lambda(\boldsymbol{z}(t))$, $p(\boldsymbol{x}|\boldsymbol{z}(t))$, $f_\theta(\cdot)$ and $\nu_\theta(\cdot)$) parameterized by neural networks does not rely on a VAE-based training pipeline. Instead, the authors leverage the adjoint method developed in \cite{chen2018neural}, with special attention given to the jump term. Interested readers are referred to the original paper for further details. 

In \cite{deng2021continuous}, the decoding from the latent states $\boldsymbol{z}_t$ to the observable $\boldsymbol{x}_t$ leverages continuously-indexed normalizing flows\footnote{The main idea of a normalizing flow is to express the joint distribution of some (generic) $\boldsymbol{y}$ as the transformation of some (generic) $\boldsymbol{x}$, i.e., $\boldsymbol{y}=T(\boldsymbol{x})$ with $\boldsymbol{x}\sim p_\text{base}(\boldsymbol{x})$ coming from some simple base distribution (e.g., multivariate normal); $T$ is invertible and differentiable, and usually comprises a sequence of transformations $T=T_K\circ  \cdots \circ T_1$.}\citep{papamakarios2021normalizing}
modeled as follows:
\begin{equation*}
    \boldsymbol{x}_t = G_\theta(\mathbf{o}_t, \boldsymbol{z}_t, t),
\end{equation*}
where $\mathbf{o}_t$ is a $k$-dimensional ``base" process with closed-form transition density (e.g., a Wiener process or an Ornstein–Uhlenbeck (OU) process), and $G_\theta(\cdot,\boldsymbol{z}_t,t)$ is a normalizing flow \citep[e.g.,][]{kobyzev2020normalizing} that transforms the base process into the more complex one through invertible mappings. The dynamics of $\boldsymbol{z}_t$ are modeled as an SDE similar to that in \cite{hasan2021identifying}. 
It is worth noting that the starting point of \cite{deng2021continuous} differs slightly from the other works reviewed thus far: from a modeling perspective, instead of treating the latent process as the governing component of the system that drives the evolution of the observable process, the latent process is incorporated primarily for expressiveness.
Specifically, built upon \cite{deng2020modeling} where the process is modeled as $\boldsymbol{x}_t = G_\theta(\mathbf{o}_t, t)$---which omits the latent process and only includes the base process, the additional inclusion of the latent process $\boldsymbol{z}_t$ enhances the model's representation capacity, thereby potentially reducing the inductive bias. At the conceptual level, connections can be drawn to classical statistical modeling where a factor-augmented structure is leveraged \citep[e.g.,][]{bernanke2005measuring,bai2016estimation,lin2020regularized}.

\subsection{Comparative remarks}

The following remarks focus on the VAE-based models reviewed above. We compare their formulation, encoding and decoding procedures, computational characteristics, and implementation.

\begin{remark}[Formulation-level comparison]
At the formulation level, the main distinction lies in how latent dynamics are specified. In latent neural ODE/SDE models, the latent process is defined over continuous time and can therefore be evaluated directly at arbitrary observation timestamps, making such formulations naturally suited to irregularly sampled data (see also Section \ref{sec:irregular-ts}). By contrast, in the standard discrete-time formulation,
\begin{equation*}
\boldsymbol{z}_t = f_{\theta}(\boldsymbol{z}_{t-1}) + \boldsymbol{u}_t,
\end{equation*}
the latent transition evolves from one observation index to the next and does not explicitly encode the elapsed time between observations. Discrete-time models can nevertheless be extended to accommodate irregular sampling by incorporating the time gap as an input, for example through
\begin{equation*}
\boldsymbol{z}_t = f_{\theta}(\boldsymbol{z}_{t-1}, \Delta t) + \boldsymbol{u}_t,
\end{equation*}
as in time-aware recurrent models such as \citet{che2018recurrent}, as well as in the original formulation of~\citet{krishnan2015deepkf}. Thus, the distinction lies mainly in how the formulation accommodates non-uniform observation intervals: continuous-time models do so directly through the continuous latent dynamics, whereas discrete-time models typically do so with an explicit time-aware extension.
\end{remark}

\begin{remark}[Operationalization of encoding and decoding]
Another point of comparison lies in how encoding and decoding are carried out. In the latent neural ODE case, the encoder compresses the information in the observed trajectory $\mathbb{x}$ into the initial latent state $\boldsymbol{z}_{t_0}$ by learning the posterior distribution $q_\phi(\boldsymbol{z}_{t_0}\mid \mathbb{x})$. A sample from this distribution then serves as the initial condition of the ODE during decoding, and the latent trajectory $\boldsymbol{z}_{t_i}$ is obtained deterministically by solving the corresponding initial value problem. By contrast, in discrete-time neural SSMs and latent neural SDE models, the encoder approximates the posterior distribution of the latent state sequence. In the latent SDE case, encoding produces approximate posterior distributions for the latent states at each $t_i$, and decoding propagates sampled latent states forward through the state transition equation (see Exhibit~\ref{sdepipe}). Discrete-time neural SSMs follow the same broad step-by-step pattern. The main difference between the latter two lies in how the transition dynamics are parameterized: discrete-time models typically use a general latent transition map, whereas latent SDE models posit an additive drift--diffusion structure, so that the latent update is expressed through an increment determined by the underlying SDE. This distinction also affects the ELBO terms used in training: in stepwise models such as discrete-time neural SSMs and latent neural SDEs, KL contributions typically arise along the latent state sequence, whereas in latent neural ODE models the KL term is commonly associated only with the inferred initial latent state, since the subsequent decoding trajectory is deterministic.
\end{remark}

\begin{remark}[Computational aspects]
A third point of comparison concerns the computational implications of these modeling choices. In discrete-time SSMs, latent states are propagated through autoregressive transitions from one observation step to the next. By contrast, continuous-time models rely on numerical solvers to compute the latent trajectory. In latent neural ODE models, this usually takes the form of solving an ODE initial value problem, often with adaptive solvers whose cost depends on the number of internal function evaluations required by the learned dynamics. Latent neural SDE models follow the same general principle, in that decoding also requires numerical discretization of the underlying continuous-time dynamics. In practice, however, their computational cost is often more implementation dependent. Simplified schemes such as one-step Euler-Maruyama updates can make the resulting computation much closer to that of a discrete-time autoregressive transition. Under such an implementation \citep[e.g.,][]{hasan2021identifying}, the latent SDE update is operationally comparable to a step-by-step discrete-time transition. Thus, the practical computational gap is not always simply between discrete-time and continuous-time formulations, but also depends on how the latter are numerically implemented.
\end{remark}

\begin{remark}[Implementation and training considerations]
The specific neural architectures used in these models are largely secondary to the underlying state-space formulation. In practice, standard feedforward networks are often sufficient for static mappings, such as those used to parameterize the emission distribution or a transition function evaluated at a given time step, whereas recurrent architectures such as GRUs or LSTMs are more naturally used in inference networks that process the observed trajectory sequentially. More generally, practical performance is often shaped by implementation and optimization choices, including latent-state dimension, KL annealing, learning-rate scheduling, the balance between reconstruction and regularization terms, and the tractable parameterization used in ELBO evaluation. In practice, the KL terms are often made computationally tractable by parameterizing both the variational posterior and the prior in Gaussian form, sometimes with further simplifications such as diagonal covariance. Although VAE-based models may in principle suffer from posterior collapse, this issue is often less pronounced in time-series settings because sequential transition dynamics provide a persistent learning signal across time.
\end{remark}

\begin{remark}[Comparison with classical inference]
Under the classical framework, learning latent continuous-time SDE models generally requires computing the continuous-time analogues of posterior distributions, such as $p(\boldsymbol{z}_t \mid \mathbb{x}_{0:s})$ arising in filtering $(s \le t)$ and smoothing $(s=T)$, which is typically a highly involved task under EM-based inference; see Appendix~\ref{sec:ct-SSM}. Recent developments have, to a large extent, alleviated this difficulty through plug-and-play methods—such as those implemented in the \texttt{pomp R} package \citep{king2016statistical}—that require only forward simulation rather than closed-form transition densities, alongside particle-based likelihood methods such as iterated filtering \citep{ionides2011iterated}, Iterated Block Particle Filters \citep{ning2023iterated}, and Particle MCMC \citep{kantas2015on}. While these approaches can provide exact or asymptotically consistent inference, they remain simulation-intensive and often require a large number of particles for each parameter update.

By contrast, the deep learning framework adopts a VAE-based approach centered on amortized inference. Rather than repeatedly approximating the continuous-time posterior through simulation-based filtering or smoothing, it uses an inference network to approximate the relevant latent distributions directly; see~\eqref{eqn:approx-SDE-encoder-1} and~\eqref{eqn:approx-SDE-encoder-2}. This replaces repeated model-based posterior computation inside the training loop by a learned encoder-based approximation, which can offer substantial computational advantages in large-scale or real-time settings. Nevertheless, such computational gains come at the cost of variational approximation and amortization errors, with the quality of posterior inference potentially sensitive to the choice of variational family, encoder architecture, and training objective.
\end{remark}

\subsection{Additional related approaches}\label{sec:early-DL-SSM}

We highlight selected works that, while adhering closely to classical approaches, leverage deep learning tools---such as neural networks for function parameterization---to enable more flexible model specification.

An example of this approach is detailed in \cite{rangapuram2018deep}. Let $\boldsymbol{x}^{j}_t\in\mathbb{R}$ be the $j$-th coordinate of the observation process, whose corresponding noise process is given by $\boldsymbol{\epsilon}^j_t$. With a slight abuse of notation, let $\boldsymbol{z}_t^{(j)}$ be the latent state and $\boldsymbol{c}^{(j)}_t$ the covariates associated with $\boldsymbol{x}^{j}_t$ --both potentially multivariate. A discrete-time linear system of the following form is considered:
\begin{align*}
\boldsymbol{z}^{(j)}_0 &\sim \mathcal{N}(\boldsymbol{\mu}_0, \Sigma_0),~~\Sigma_0\text{~is diagonal},\\
\boldsymbol{z}^{(j)}_{t} &= F^{(j)}_t\boldsymbol{z}^{(j)}_{t-1} + G^{(j)}_t \boldsymbol{u}^{(j)}_t,\\
\boldsymbol{x}^{j}_t &= b^{(j)}_t + H^{(j)}_{t}\boldsymbol{z}^{(j)}_t + \boldsymbol{\epsilon}^{j}_t,~~ \boldsymbol{\epsilon}^{j}_t\sim\mathcal{N}(\boldsymbol{0},r^{(j)}_t),
\end{align*}
with time-varying parameters 
\begin{equation*}
    \theta^{(j)}_t:=(\boldsymbol{\mu}_0, \Sigma_0, F^{(j)}_t, G^{(j)}_t, b^{(j)}_t, H^{(j)}_t, r^{(j)}_t),
\end{equation*}
parameterized through $\theta_t^{(j)}=f_{\phi}(\mathbb{c}^{(j)}_{1:t})$. The mapping $f_{\phi}$ is shared across all $j$'s and is parameterized by an RNN, and the parameter $\phi$ is learned by maximizing the log-likelihood of the observations. In particular, in the Gaussian case,  the likelihood can be decomposed as in~\eqref{eq77:likelihood-SSM-one-step}, and the corresponding terms can be calculated using the Kalman filter. The authors also briefly discuss using a VAE to optimize the lower bound, when the Gaussian assumption does not hold; see Appendix A of the paper for details. Notably, the time-varying parameters are handled by neural networks, which conveniently provide the necessary mapping for each time step as a result of the inner workings of RNNs. A similar idea is adopted in~\cite{revach2022kalmannet}, where an RNN is used to parameterize the Kalman gain to perform recursive filtering for the real-time state estimation problem.

\begin{remark}[Connection to time-varying linear Gaussian SSMs] 
The model presented above is a \textit{time-varying linear} Gaussian SSM whose parameters are specified by neural networks. Classical approaches often impose additional structure on the time-varying parameters to make estimation tractable. For example, a switching SSM uses regime-specific coefficient matrices governed by a finite-state Markov chain (see Appendix~\ref{sec:discrete-linear-Gaussian-SSM}). The coefficients are piecewise constant, while the state dynamics are linear within each regime. Neural networks provide a flexible alternative for specifying the time variation.
\end{remark}

In \cite{masti2021learning}, a non-linear state-space model is proposed in the form\footnote{For simplicity, the control variable from the original model is omitted}
\begin{equation*}
    \boldsymbol{z}_{t+1} = f(\boldsymbol{z}_{t}, \boldsymbol{u}_{t+1}), \qquad \boldsymbol{x}_{t} = g(\boldsymbol{z}_{t}, \boldsymbol{\epsilon}_t). 
\end{equation*}
At the core of the proposed framework, the problem is recast as solving for $f$ and $e,d$, where $\boldsymbol{z}_t := e(\mathbb{x}_{t-k:t-1}),k>1$, with $e(\cdot)$ being the ``encoder" that performs dimensionality reduction, while the decoder  $d(\cdot)$ reconstructs $\mathbb{x}_{t-k':t}, k'<k$\footnote{In the original paper, given the presence of the control variable, the input to the encoder encompasses also past information of the control variable, whereas the reconstruction target always only encompasses the observation $\boldsymbol{x}_t$'s.}. Here, $f$ handles state transitions and propagates information from $t$ to $t+1$. The functions $f,e,d$ are jointly learned by minimizing the loss that comprises two components: one for reconstruction error based on observed data, and another based on the discrepancy between the state estimates respectively from state transitions and from the encoder, for the same time point.

A similar approach within the scope of factor models can be found in the deep dynamic factor model by \cite{andreini2020deep}, where factor (or state) dynamics are modeled as a linear VAR process. Related work on deep factor models is also explored in \cite{pmlr-v97-wang19k}.

%% file: 04-lssl-s4-mamba.tex
\section{SSMs as Components of an Input-Output System}\label{sec:semi-implicit}

In this section, we diverge from the discussion in previous sections and consider the body of work where SSMs are utilized as input-output mapping modules. 

The advent of Transformers \citep{vaswani2017attention} led to the development of language models with enhanced capabilities, including BERT \citep{devlin2018bert}, T5 \citep{T5} and the GPT family \citep{radford2018improving, GPT3}. However, these models suffer from computational complexity that scales quadratically with the length of the input sequence, which becomes a severe bottleneck when processing long sequences. In particular, self-attention requires forming pairwise interactions across all tokens, leading to $\mathcal{O}(L^2)$ time and memory complexity.

In response to this challenge, various alternatives to traditional sequence modeling paradigms such as CNNs, RNNs, and Transformers have emerged \citep[e.g.,][]{dai2019transformer,kitaev2020reformer}, though they still face computational and learning-related limitations.

Recent works, such as those in \cite{gu2021combining,gu2022efficiently,smith2023simplified,gu2023mamba,dao2024transformers}, have focused on enhancing long-range sequence modeling performance across a range of application domains. These recent SSM-based approaches can overcome some of the bottlenecks of previous models, and excel in modeling tasks involving very long sequences. 

At a high level, the sequence modeling task can be generically cast as establishing a mapping from the input sequence $\{\boldsymbol{x}_{t-L+1},\cdots,\boldsymbol{x}_t\}$ with context length $L$ to the output sequence $\{\boldsymbol{x}_{t+1},\cdots,\boldsymbol{x}_{t+L'}\}$ of length $L'$. For the line of work in question, by treating the SSMs as input-output systems and incorporating them into neural network architectures, efficient processing of long input sequences becomes achievable. This improvement arises from replacing explicit pairwise interactions with structured state propagation, leading to linear-time complexity in sequence length. In particular, these models admit convolutional or recurrent (scan-based) implementations, the latter enabling streaming inference with constant memory, as only the current state is maintained. Long-range dependencies are captured implicitly through the state evolution rather than explicit token-to-token interactions.

In contrast to the work reviewed in Sections~\ref{sec:intro-SSM} and~\ref{sec:neural-SSM}, where the models are postulated as having an inherent latent component, here SSMs are not associated with specific latent state-driven DGPs. Instead, they serve as stand-alone input-output mapping modules that propagate information. This approach is implemented with Linear State-Space Layers (LSSLs) as the operating unit, and the technical development focuses on the efficient computation and learning of these SSM-enabled modules. 

\subsection{The main building block: linear state-space layers (LSSLs)}

\cite{gu2021combining} introduce LSSLs as a new paradigm that integrates popular tools from sequence modeling, including recurrence, convolution and differential equations. 

Concretely, let $\boldsymbol{u}(t)\in\mathbb{R}^{d_{\text{in}}}$ denote the input\footnote{Throughout this section, $\boldsymbol{u}(t)$ represents the input signal to the SSM module, rather than the process noise used in Sections~\ref{sec:intro-SSM} and \ref{sec:neural-SSM}.}, $\boldsymbol{v}(t)\in\mathbb{R}^{d_{\text{out}}}$ the output and $\boldsymbol{z}(t)$ the $d$-dimensional latent state. A continuous-time linear state-space model is defined by
\begin{equation}\label{eqn:LSSL-cont}
\frac{\dd \boldsymbol{z}(t)}{\dd t} = A\boldsymbol{z}(t) + B\boldsymbol{u}(t),~~\boldsymbol{v}(t) = C\boldsymbol{z}(t) + D\boldsymbol{u}(t).
\end{equation}
After discretization, the system in~\eqref{eqn:LSSL-cont} possesses a {\em recurrent} form given by:
\begin{equation}\label{eqn:LSSL-rec}
    \boldsymbol{z}_t = \bar{A}\boldsymbol{z}_{t-1} + \bar{B} \boldsymbol{u}_t, \qquad \boldsymbol{v}_t = C \boldsymbol{z}_t + D\boldsymbol{u}_t,
\end{equation}
where $\bar{A}$ and $\bar{B}$ are approximations of $A$ and $B$ using the bilinear method \citep{tustin1947method}, respectively:
\begin{equation}\label{eqn:bilinear}
  \bar{A}:= (I-\tfrac{\Delta}{2}A)^{-1} (I+\tfrac{\Delta}{2}A),~~ \bar{B} := (I-\tfrac{\Delta}{2}A)^{-1}\Delta B.  
\end{equation}
Here, $\Delta$ is the timescale at which the input $\boldsymbol{u}(t)$ is discretized and thus $\boldsymbol{u}_t=\boldsymbol{u}(\Delta\cdot t)$; $\boldsymbol{v}_t$ and $\boldsymbol{z}_t$ are analogously defined. For an input sequence starting at $t=0$ and zero initial state $\boldsymbol{z}_{-1}=0$, unrolling the recurrence in~\eqref{eqn:LSSL-rec} leads to
\begin{equation*}
    \boldsymbol{v}_t = C\bar{A}^t\bar{B}\boldsymbol{u}_0 + C\bar{A}^{t-1}\bar{B}\boldsymbol{u}_1 + \cdots + C\bar{B}\boldsymbol{u}_t + D\boldsymbol{u}_t.
\end{equation*}
In particular, for a discrete input sequence $\{\boldsymbol{u}_t\}$ of length $L$, based on the connection between a linear time-invariant system and continuous convolution\footnote{For a linear time-invariant system, let $h(t)$ be the impulse response function. The output $\boldsymbol{v}(t)$ in response to the input $\boldsymbol{u}(t)$ then satisfies $\boldsymbol{v}(t)\approx\sum_j h(t-\tau_j)\boldsymbol{u}(\tau_j)\Delta\tau$; taking the mesh width $\Delta\tau\rightarrow 0$, $\boldsymbol{v}(t)=\int_{-\infty}^\infty h(t-\tau)\boldsymbol{u}(\tau)\dd\tau$, i.e., $\boldsymbol{v}(t)=h(t) * \boldsymbol{u}(t)$. For further details, see \cite{oppenheim1983signals}. }, the following holds with an SSM convolution kernel $\bar{K}$ and convolution operator $*$:
\begin{align}
&\boldsymbol{v} = \bar{K} * \boldsymbol{u} + D\boldsymbol{u},\label{eqn:LSSL-conv} \\
\text{where}~~&\bar{K}\in\mathbb{R}^L:=\mathcal{K}_L(\bar{A},\bar{B},C) := (C\bar{B},C\bar{A}\bar{B},\cdots,C\bar{A}^{L-1}\bar{B}).\nonumber
\end{align}
The representation in~\eqref{eqn:LSSL-conv} corresponds to the {\em convolutional} form, which can be computed efficiently using the fast Fourier transform (FFT). 

Equations~\eqref{eqn:LSSL-cont},\eqref{eqn:LSSL-rec} and~\eqref{eqn:LSSL-conv} represent different perspectives of an LSSL, each contributing to its versatility and enabling this paradigm to leverage a combination of advantages. Specifically, as with many other continuous-time models, LSSLs naturally adapt to different timescales, and their connection to differential equations ensures that the trajectory is mathematically tractable. On the other hand, being a recurrent model, they are ``stateful" and hence allow for fast autoregressive generation. Finally, the convolutional representation enables parallelizable training. The authors further establish explicit connections between LSSLs and convolutions, with the timescale $\Delta$ viewed as controlling the width of the convolution kernel. A similar connection can be established with RNNs, where the gating heuristics are linked to $\Delta$ and can be derived from ODE approximations. 

The trainable parameters in an LSSL are $A,B,C,D$, and the timescale $\Delta\in\mathbb{R}$. When both $\boldsymbol{u}(t)$ and $\boldsymbol{v}(t)$ are 1-dimensional, an LSSL effectively forms a sequence to sequence map $\mathbb{R}^L\mapsto\mathbb{R}^L$ as a single-input-single-output (SISO) system, with $B\in\mathbb{R}^{d\times 1}, C\in\mathbb{R}^{1\times d}$ and $D\in\mathbb{R}^{1\times 1}$. Multi-dimensional outputs can be easily accommodated with $C\in\mathbb{R}^{d_{\text{out}}\times d}$, $D\in\mathbb{R}^{d_{\text{out}}\times 1}$.  When the input is multi-dimensional with $d_{\text{in}}>1$, multiple copies of LSSLs are used to process each input dimension independently. Finally, LSSLs can be stacked to form deep LSSLs, akin to the modules in Transformers.

It is worth noting that the basic form of LSSLs does not perform particularly well and inherits some of the challenges of recurrence and convolution, such as vanishing gradients. To overcome this, the authors leverage the HiPPO framework \citep{gu2020hippo} that introduces continuous-time memorization. Specifically, by utilizing a class of structured matrices $A$\footnote{Known as the ``HiPPO" matrix.  Specifically, for $A\in\mathbb{R}^{d\times d}$, a HiPPO-LegS (short for ``scaled Legendre measure") matrix is given by $A_{ij}=-(2i+1)^{1/2}(2j+1)^{1/2}$ if $i>j$; $-(i+1)$ if $i=j$ and $0$ otherwise.\label{foot:hippo}}, the state $\boldsymbol{z}(t)$ can effectively ``memorize" the history of the input $\boldsymbol{u}(t)$. This approach significantly enhances the LSSL's performance, particularly in tasks involving long-sequence modeling. 

In summary, an LSSL is a sequence modeling module, whose design primarily concerns 1-dimensional sequences. It maps an input sequence $\boldsymbol{u}(t)$ to the output sequence $\boldsymbol{v}(t)$ via the latent state $\boldsymbol{z}(t)$, by emulating a linear continuous-time SSM representation in discrete time. Multi-dimensional input/output can be handled with necessary modifications/adaptations.

\subsection{Structured state-spaces (S4)}

Despite its appealing properties, the computation of LSSLs remains a bottleneck in practice. As pointed out in \cite{gu2022efficiently}, with state dimension $d$ and context length $L$, computing the latent state requires $O(d^2L)$ operations, while memory usage scales as $O(dL)$. In addition, although a theoretically-efficient algorithm exists, it is not numerically stable. The special form of the HiPPO matrix $A$ restricts the use of convolution techniques. \cite{gu2022efficiently} address these limitations and enhance the practical applicability of LSSLs, focusing on improving  both memory and computational efficiency. The primary setting considered involves both the input $\boldsymbol{u}(t)$ and the output $\boldsymbol{v}(t)$ being 1-dimensional, i.e., a single-input, single-output (SISO) SSM. In addition, $D$, see ~\eqref{eqn:LSSL-cont} and~\eqref{eqn:LSSL-rec}, is omitted hereafter (or equivalently, assume $D=0$), due to the fact that the term $D\boldsymbol{u}$ can be viewed as a skip connection and thus is easy to compute.

The major technical advancement involves considering the generating function on the coefficients of the convolution kernel $\bar{K}$ and imposing a special structure on the state matrix $A$. These two elements together enable stable and efficient computation, in the steps where repeated matrix multiplication of $\bar{A}$ is required (see, e.g., equation~\eqref{eqn:LSSL-conv}). Concretely, the key steps can be outlined as follows:
\begin{enumerate}[itemsep=0pt]
\setlength{\leftmargini}{0pt}
\item Instead of directly computing $\bar{K}$, its spectrum is computed through the truncated generating function 
$\widetilde{\mathcal{K}}_L(\phi; \bar{A},\bar{B},C) := \sum\nolimits_{\ell=0}^{L-1} C\bar{A}^{\ell}\bar{B} \phi^\ell$ evaluated at the roots of unity $\phi:=\{\exp(2\pi i \ell/L): \ell=1,\cdots,L\}$. The filter $\bar{K}$ is then recovered by applying the inverse FFT in $O(L\log L)$ operations. Crucially, using the properties of the truncated generating function, that is, by letting $\widetilde{C}$ collect the constant terms, we can express it as\footnote{this is due to $\widetilde{\mathcal{K}}_L(\phi; \bar{A},\bar{B},C) =  \sum\nolimits_{\ell=0}^{L-1} C\bar{A}^{\ell}\bar{B} \phi^\ell = C(I-\bar{A}^L\phi^L)(I-\bar{A}\phi)^{-1}\bar{B} = \widetilde{C}(I-\bar{A}\phi)^{-1}\bar{B}$; the last equality holds since $\phi^L=1$ by definition and $\widetilde{C}$ collects the constant term}
\begin{align}\label{eqn:truncate}
  \widetilde{\mathcal{K}}_L(\phi;&\bar{A},\bar{B},C) = \widetilde{C}(I-\bar{A}\phi)^{-1}\bar{B} \stackrel{\text{discretize}}{=}\frac{2\Delta}{ 1 + \phi}\widetilde{C}\big[2\frac{1-\phi}{1+\phi}I -\Delta A\big]^{-1}B.
\end{align}
This approach involves matrix inversion rather than calculating a matrix power series, and thus it reduces to obtaining the inverse efficiently. 
\item Let the state matrix $A$ be a structured one, in the form of the sum of diagonal-plus-low-rank (DPLR) matrices, that is $A:=\Lambda -PQ^*$ with $\Lambda$ being diagonal, $P,Q\in\mathbb{R}^{d\times r}$ for some small $r$. Using the Woodbury identity \citep{woodbury1950inverting}, for $r=1$, $P,Q$ become vectors and~\eqref{eqn:truncate} satisfies
\begin{align*}
    \widetilde{K}_{L}(\phi; \bar{A},\bar{B},C) = \frac{2}{1+\phi}\Big[\widetilde{C} \kappa_\Lambda(\phi) B - \widetilde{C}\kappa_\Lambda(\phi) P
    \big[ 1+ Q^*\kappa_\Lambda(\phi) P\big]^{-1}Q^*\kappa_\Lambda(\phi) B \Big],
\end{align*}
where $\kappa_\Lambda(\phi) := (\tfrac{2}{\Delta}\frac{1-\phi}{1+\phi}I-\Lambda)^{-1}$ \citep[][Lemma C.3]{gu2022efficiently}. This reduces the calculation to $\tilde{O}(d+L)$ operations and $O(d+L)$ space, as $Q^*\kappa_\Lambda(\phi)P$ can be handled with Cauchy matrix-vector multiplication \citep[][Appendix C.3, Proposition 5]{gu2022efficiently}.
\end{enumerate}
The above techniques allow for the efficient calculation of the convolution filter $\bar{K}$. For the recurrence in~\eqref{eqn:LSSL-rec}, it can be computed in $O(d)$ operations with a DPLR matrix. By considering matrix $A$ in DPLR form, S4 is (near)-optimal for both the recurrent \eqref{eqn:LSSL-rec} and the convolutional \eqref{eqn:LSSL-conv} representations of the LSSLs. Depending on the exact setting, the former is typically used for online generation, while the latter is used for offline training with equally-spaced observations, taking advantage of the parallelizable nature of the FFT.

\begin{remark}[Some additional results]\label{rmk:S4} We briefly outline some other results established in \cite{gu2022efficiently}, focusing on the motivation behind adopting a DPLR form for the state matrix $A$.  
\begin{itemize}
    \item Conjugation is an equivalence relation for SSMs \citep[Lemma 3.1]{gu2022efficiently}. Two SSMs that are conjugate to each other provide the same $\boldsymbol{u}\mapsto\boldsymbol{v}$ mapping, but with a change of basis in the state $\boldsymbol{z}$, effectively allowing the system to be ``reparametrized". The HiPPO matrices \citep{gu2020hippo,gu2021combining} have a normal\footnote{A complex square matrix $A$ is normal if it commutes with its conjugate transpose $A^*$.}-plus-low-rank (NPLR) form (as shown in Theorem 1), and can be conjugated into a DPLR form. This has two important implications: (1) from a computation perspective, it suffices to focus on the complexities of the SSMs whose $A$ is in DPLR form; (2) practically, the SSM can be initialized with $A$ set to a specific HiPPO matrix (e.g., the HiPPO-LegS shown in footnote~\ref{foot:hippo}), with $\Lambda,P,Q$ being the actual learnable parameters associated with it.
    \item When $A$ is diagonal, the computation of the generating function reduces to calculating a Cauchy kernel, which is a well-studied problem with stable numerical algorithms (see Appendix C.3, Proposition 5). The DPLR form can be viewed as a relaxation of the diagonal case, adopted because the diagonalization of the HiPPO matrix is unstable.  This instability makes the naive application of a diagonal parameterization prone to numerical issues, thus necessitating the DPLR approach.
\end{itemize}
\end{remark}

\subsection{Simplified state-spaces (S5)}

\cite{smith2023simplified} extend S4 along the following two directions: (1) in the case where the input/output are multi-dimensional, instead of using multiple SISO SSMs followed by a ``mixing layer" to combine the features, a multi-input-multi-output (MIMO) SSM is used; this reduces the combined latent state size $(d_{\text{in}}\cdot d)$ to some $\tilde{d}$, which can be much smaller; and (2) instead of taking the frequency domain approach with generating functions, S5 utilizes parallel scan \citep{blelloch1990prefix} which yields the same computational complexity as S4, yet operates solely in the time domain via recurrence. At a high level, a parallel scan computes the linear recurrence in~\eqref{eqn:LSSL-rec} in parallel via the following steps:
\begin{enumerate}[itemsep=0pt]
\setlength{\leftmargini}{0pt}
    \item Precompute $c_k :=(c_{k,a},c_{k,b}) :=(\bar{A},\bar{B}\boldsymbol{u}_k)$ for $k=1,\cdots,L$, and let $r_0=(I,0)$. Define a binary associative operator $\bullet$ used for combining the elements $q_i\bullet q_j=(q_{j,a}\odot q_{i,a}, q_{j,a}\otimes q_{i,b}+q_{j,b})$, where $\odot,\otimes,+$ are matrix-matrix multiplication, matrix-vector multiplication and elementwise addition, respectively. 
    \item A reduction (or upsweep) step giving rise to the ``upsweep tree". Let $c_k$ be the leaves; each element is obtained by pairwise operations as defined by the operator. This step gives the partial ``prefix" scan results.
    \item A downsweep step that completes the scan. Let $r_0$ be the root. Based on the above outputs, complete the scan by passing each node's own value to the left child; the right child is obtained by applying the operator between the node itself and the left child in the (mirrored) upsweep tree.
\end{enumerate}
Given sufficient number of processors, the parallel time scales logarithmically with $L$ \citep{smith2023simplified}; see also Appendix H therein for a concrete example for a sequence of length $4$. 

\paragraph{Diagonal state space.} Similar to S4 where the efficient computation requires a specific parameterization of $A$ (e.g., a DPLR form), S5 parameterizes $A$ as a diagonal matrix\footnote{The diagonal parameterization makes the scan operator cheaper to evaluate.}. Initialization of the state matrix is done with the HiPPO-N matrix, that is, the normal part of the HiPPO-LegS matrix (see footnote~\ref{foot:hippo}). Recall that all HiPPO matrices have an NPLR form \citep{gu2022efficiently}; specifically, in the case of HiPPO-LegS,
it has the following decomposition:
\begin{align*}
A_{\text{LegS},ij}
&:= -\begin{cases}
(2i+1)^{\tfrac{1}{2}}(2j+1)^{\tfrac{1}{2}}, & i>j,\\
i+1, & i=j,\\
0, & \text{otherwise},
\end{cases}\\
A_{\text{LegS}}
&= A^{\text{Normal}}_{\text{LegS}}
   - P_{\text{LegS}}P_{\text{LegS}}^\top,
\end{align*}
where $P_{\text{LegS}}\in\mathbb{R}^{d\times 1}$ is rank-1:
\begin{align*}
P_{\text{LegS},i} := (i+\tfrac{1}{2})^{\tfrac{1}{2}},\qquad
\text{and}\quad &A^{\text{Normal}}_{\text{LegS},ij} := -\begin{cases} (i+\tfrac{1}{2})^{\tfrac{1}{2}}(j+\tfrac{1}{2})^{\tfrac{1}{2}}~~&i>j, \\\tfrac{1}{2}&i=j,\\ -(i+\tfrac{1}{2})^{\tfrac{1}{2}}(j+\tfrac{1}{2})^{\tfrac{1}{2}}~~&i<j;\\ \end{cases}.
\end{align*}
This normal portion $A^{\text{Normal}}_{\text{LegS}}$ can be diagonalized
stably \citep{gupta2022diagonal, gu2022parameterization}, though
diagonalization of the full matrix $A_{\text{LegS}}$ suffers from some instability issues.

\subsection{Selective-scan state-spaces (Mamba)}

In the SSM used thus far for mapping the input $\boldsymbol{u}(t)$ to the output $\boldsymbol{v}(t)$, the model parameters (e.g., $A,B,C,D$) are time-invariant. However, \cite{gu2023mamba} argue that the invariance in the dynamics restricts the model's ability to ``selectively" incorporate contextual information, limiting its capacity to update hidden states based on input-specific cues. To address this, an SSM with a selection mechanism is proposed, called Mamba, where the parameters are functions of the input and~\eqref{eqn:LSSL-rec} is extended to the following form:
\begin{equation*}
\boldsymbol{z}_t = \bar{A}(\boldsymbol{u}_t)\boldsymbol{z}_{t-1} + \bar{B}(\boldsymbol{u}_t) \boldsymbol{u}_t, \qquad \boldsymbol{v}_t = C(\boldsymbol{u}_t) \boldsymbol{z}_t.
\end{equation*}
Notably, instead of having fixed $\bar{A}, \bar{B}, C$, they are now functions of $\boldsymbol{u}_t$ and thus become context-aware, and the SSM now corresponds to a time-varying system. Following the discretization rule in~\eqref{eqn:bilinear} or its zero-order hold (ZOH) alternative, given by
\begin{equation}\label{eqn:ZOH}
    \bar{A} := \exp(\Delta A), ~~\bar{B} := (\Delta A)^{-1}\big(\exp(\Delta A) - I\big) \cdot \Delta B,
\end{equation}
the input-dependent state transition matrix $\bar{A}(\boldsymbol{u}_t)$ is operationalized ``indirectly" via the discretization time scale $\Delta$ that becomes a function of $\boldsymbol{u}_t$, given in the form of $\Delta(\boldsymbol{u}_t) := \text{softplus}\big(c+\text{Linear}(\boldsymbol{u}_t)\big)$\footnote{$\text{softplus}(x):=\log\big(1+\exp(x)\big)$. In the expression, $c$ is some (hyper)-parameter; we also omit the ``$\text{broadcast}_{D}$" operation in the original paper (Algorithm 2), which effectively creates identical $D$ copies along the ``extra" dimension so that the tensor shape is suitable for subsequent operations.}. Meanwhile, $A$ remains an input-independent structured matrix, while $B$ and $C$ are assumed linear functions of $\boldsymbol{u}_t$ that directly govern the magnitude of the input $\boldsymbol{u}_t$'s influence on the latent state $\boldsymbol{z}_t$, and that of the state $\boldsymbol{z}_t$ on the output $\boldsymbol{v}_t$, respectively. In the special case where the state dimension is 1, connections between the selective SSM recurrence and the gating mechanism used in RNN can be further established. Further, $\Delta$ can be interpreted as controlling the amount of emphasis on the current input $\boldsymbol{u}_t$: a small $\Delta$ corresponds to the case where the current input is considered transient and thus ``unselected", whereas a larger $\Delta$ places more emphasis on the current input, rather than the context that is contained in the state. Similar to S5, Mamba leverages parallel scanning and emphasizes the recurrence perspective of the SSM, with the state matrix $A$ assumed to have a diagonal structure. 
 
\begin{remark}
To compute selective state spaces efficiently with parallel scan, the authors consider additional techniques that fully leverage modern GPU capabilities. These include using kernel fusion to reduce memory I/O operations, performing discretization and recurrence calculations directly in GPU SRAM while loading the ``raw" parameters from GPU HBM\footnote{SRAM and HBM are shorthand for "static random-access memory" and ``high-bandwidth memory", respectively.}, and leveraging recomputation to avoid saving intermediate states needed for backpropagation, thereby reducing memory requirements. The specific implementation details are omitted here; readers interested in a comprehensive explanation are encouraged to consult the original paper.
\end{remark}

\medskip
To conclude this section, it is worth noting that modern structured SSMs represent an active area of research in the deep learning community. On the theoretical front, \cite{cirone2024theoretical} propose a continuous-time framework that establishes connections between Linear Controlled Differential equations (Linear CDEs) with SSMs, such as S4 and Mamba. The framework provides an explicit characterization of the ``uniform closure"\footnote{The uniform closure describes the set of functions from compact subsets of the input space to $\mathbb{R}$ (the output space) that can be uniformly approximated at an arbitrary precision by a Linear CDE (see Section 4 of \cite{cirone2024theoretical})}, along with insights into the efficiency, stability, and expressiveness of these systems in relation to their specific structured transition matrices or architectures. Empirically, SSM-based architectures such as Linear Attention \citep{katharopoulos2020transformers} that can be viewed as a degenerate linear SSM \citep{gu2023mamba}, and more recent models such as RetNet \citep{sun2023retentive} have gained popularity as efficient alternatives to standard Transformers that suffer from quadratic computational complexity. Additional connections between Transformers and SSMs are explored in the follow-up work Mamba-2 \citep{dao2024transformers}. S4/S5/Mamba-style architectures have also found applications across other areas, including spatio-temporal modeling \citep{smith2024convolutional}, vision \citep{zhu2024vision} and graph learning \citep{wang2024graph}.

%% file: 05-tasks.tex
\section{Applications to Selected Modeling Tasks}
\label{sec:applications}

In this section, we first present two key time series modeling tasks facilitated by the SSM formulation: nowcasting and the imputation of irregularly spaced data. We then provide a brief overview of how the previously discussed deep learning-based SSMs are applied, focusing specifically on scientific domains where the underlying dynamics are captured by mechanistic models.

\subsection{Nowcasting with neural state space models}\label{sec:nowcasting-nSSM}

Nowcasting concerns estimating the current state of a partially observed dynamical system, where at any given time only a subset of the variables are observed due to asynchronous data releases or inherent missingness. By leveraging the available observations to infer the underlying latent state, nowcasting effectively reconstructs the full system at the current time, enabling downstream tasks such as forecasting and decision-making.  

Classical approaches to nowcasting are largely based on linear SSMs, most notably dynamic factor models that provide a parsimonious representation of high-dimensional time series in mixed-frequency settings \citep{giannone2008nowcasting}. In this framework, the high-frequency variables $\boldsymbol{x}_t$ and the latent state $\boldsymbol{z}_t\in\mathbb{R}^d$ evolve as a system
\begin{equation*}
\boldsymbol{z}_t =  A \boldsymbol{z}_{t-1}+\boldsymbol{u}_t; \quad 
\boldsymbol{x}_t =  C \boldsymbol{z}_t+ \boldsymbol{\epsilon}_t.
\end{equation*}
Low-frequency variables $y_{j,t'}$, observed at a coarser time scale, are then linked to the latent state via regression equations of the form\footnote{$r$ is the frequency ratio between the high- and low-frequency variables.}
\begin{equation*}
y_{j,t'} = \alpha_j + \boldsymbol{\beta}_j^\top \boldsymbol{z}_{rt'} + \varepsilon_{j,t'},\qquad j=1,\cdots,q.
\end{equation*}

\begin{algorithm*}[!bt]
\SetAlgorithmName{Exhibit}{}{}
\captionsetup{font=small}
\small
\caption{A VAE-based Masked Training Pipeline for DL-SSM Nowcasting}\label{algo:nowcast}
\KwIn{Trajectory samples $\{(\mathbb{x}^{(i)},\mathbb{m}^{(i)})\}_{i=1}^n$, where $\mathbb{x}^{(i)}:=\{\boldsymbol{x}^{(i)}_{1},\cdots,\boldsymbol{x}^{(i)}_{L}\}$ and $\mathbb{m}^{(i)}:=\{\boldsymbol{m}^{(i)}_{1},\cdots,\boldsymbol{m}^{(i)}_{L}\}$}\vspace*{1mm}
\textbf{-- Forward pass:}  (for each sample indexed by $(i)$, and we omit the superscript for notational simplicity)\\
\Indp 
1. Encoder: pass $\mathbb{x}$ and $\mathbb{m}$ through their respective projection layers, combine the representation, then pass it to a bidirectional RNN and a final output projection module. The outputs give rise to
\begin{equation*}
    q_\phi(\boldsymbol{z}_t | \mathbb{x}, \mathbb{m}) = \mathcal{N}\big(\mu_{t,\text{enc}}(\mathbb{x}, \mathbb{m}), \sigma^2_{t,\text{enc}}(\mathbb{x},\mathbb{m})\big)~\text{for $t=1,\cdots,L$}.
\end{equation*}\\
2. Sampling: sample $\widetilde{\boldsymbol{z}}_t$'s according to the encoded distribution, using the reparameterization trick.\\
3. Creating pseudo-masks: generate a binary mask $\widetilde{\mathbb{m}}:=\{\widetilde{\boldsymbol{m}}_1,\cdots,\widetilde{\boldsymbol{m}}_L\}$. The corresponding masked trajectory is given by $\check{\mathbb{x}} := \mathbb{x}\odot\widetilde{\mathbb{m}}$, and the guidance mask is given by $\check{\mathbb{m}}:=\mathbb{m}\odot\widetilde{\mathbb{m}}=\{\check{\boldsymbol{m}}_1,\cdots,\check{\boldsymbol{m}}_L\}$.\\
4. Decoder: for all $t=2,\cdots,L$\\
\hspace*{3mm}-- state prediction: compute the unguided latent prior
\begin{equation*}
p_\theta(\boldsymbol{z}_t\mid
\boldsymbol{z}_{t-1}=\widetilde{\boldsymbol{z}}_{t-1})
=
\mathcal{N}\big(
\boldsymbol{\mu}_t^-,
\operatorname{diag}(\boldsymbol{v}_t^-)
\big),
\quad
\boldsymbol{\mu}_t^-:=f_{\mathrm{mean}}(
\widetilde{\boldsymbol{z}}_{t-1}),
\quad
\boldsymbol{v}_t^-:=f_{\mathrm{var}}(
\widetilde{\boldsymbol{z}}_{t-1});
\end{equation*}\\
\hspace*{3mm}-- state correction: update $\boldsymbol{\mu}_t^-$
using $(\check{\boldsymbol{x}}_t,\check{\boldsymbol{m}}_t)$
to obtain $\widehat{\boldsymbol{z}}_t$;\\
\hspace*{3mm}-- state-observation emission: $p_\theta(\boldsymbol{x}_t | \boldsymbol{z}_t=\widehat{\boldsymbol{z}}_t)= \mathcal{N}\big( g_{\text{mean}}(\widehat{\boldsymbol{z}}_t), g_{\text{var}}(\widehat{\boldsymbol{z}}_t)\big)$. \vspace*{1mm}\\
\Indm\textbf{-- Loss calculation} \\ 
\Indp
5. Calculate the training loss for all samples according
to~\eqref{eqn:masked-loss}, with the prediction error evaluated
only on withheld observed entries indexed by
$\{(t,j):m_{t,j}=1,\ \widetilde{m}_{t,j}=0\}$.\vspace*{1mm}\\
\Indm\textbf{-- Backward pass}: update neural network parameters based on gradients (back-propagation).\vspace*{1mm}\\
\KwOut{Trained encoder (parameter $\phi$) and decoder (parameter $\theta$)}
\end{algorithm*}
This structure allows information from the high-frequency block to inform current estimates of the low-frequency variables, with inference typically carried out through Kalman filtering and smoothing. While effective and interpretable, such linear Gaussian formulations are limited by their rigid parametric structure and can become computationally burdensome in large systems. In particular, estimating the model parameters typically entails repeatedly applying Kalman filtering/smoothing across optimization iterations, with the cost increasing substantially as the observation dimension and time horizon grow. 

Deep learning-based SSMs build on the same low-rank state space backbone, but replace rigid linear parameterizations with flexible neural dynamics and observation mappings, thereby accommodating richer nonlinear dependence while scaling more effectively to high-dimensional systems. A key challenge in this setting is enabling contemporaneous inference, i.e., updating the latent state using partially observed information at the current time step, analogous to the role of the Kalman gain in classical filtering. This can be achieved through a masked VAE-style training pipeline, where the model is trained to condition on partially observed inputs and infer the corresponding latent state, effectively learning to incorporate guidance information at inference time. 

Formally, using the notation from Section~\ref{sec:neural-SSM}, consider the discrete-time deep SSM
\begin{equation*}
\boldsymbol{z}_t = f(\boldsymbol{z}_{t-1})+\boldsymbol{u}_t,\qquad
\boldsymbol{x}_t = g(\boldsymbol{z}_t)+\boldsymbol{\epsilon}_t,
\end{equation*}
where $f$ and $g$ are parameterized by neural networks. Importantly, the underlying data-generating formulation remains the same as the generic discrete-time SSM introduced earlier; the nowcasting-specific issue is not a change to the state equation itself, but rather an inference requirement. Namely, at test time, the model must be able to update its estimate of the current latent state $\boldsymbol{z}_t$ using not only information propagated from the past through $\boldsymbol{z}_{t-1}$, but also the subset of coordinates of $\boldsymbol{x}_t$ that are contemporaneously observed. To describe the masked training procedure, let $\mathbb{x}:=\{\boldsymbol{x}_1,\cdots,\boldsymbol{x}_L\}$ denote a trajectory sample of length $L$, and let $\mathbb{m}:=\{\boldsymbol{m}_1,\cdots,\boldsymbol{m}_L\}$ denote the corresponding observation masks. Endowing the model with the required inference capability then calls for a masked training pipeline, summarized in Exhibit~\ref{algo:nowcast}.

Specifically, for each training sample $(\mathbb{x}^{(i)},\mathbb{m}^{(i)})$, an additional pseudo-mask
$\widetilde{\mathbb{m}}$ is introduced on the observed entries to emulate the partial-information pattern expected at inference time. The encoder takes both $\mathbb{x}$ and $\mathbb{m}$ as input---implemented in the reference setup via projection layers followed by a bidirectional RNN---and produces Gaussian latent distributions $q_\phi(\boldsymbol{z}_t\mid\mathbb{x},\mathbb{m})$, from which latent samples are drawn using the reparameterization trick. The decoder first propagates the state through the unguided
transition $p_\theta(\boldsymbol{z}_t\mid\widetilde{\boldsymbol{z}}_{t-1})$. Given the pseudo-masked input
$(\check{\boldsymbol{x}}_t,\check{\boldsymbol{m}}_t)$, a learned correction then updates the predicted latent mean using the currently observed coordinates. This updated state is mapped through the observation equation to reconstruct $\boldsymbol{x}_t$, with the reconstruction error evaluated only on the withheld observed entries. Training combines prediction error on withheld observed entries with a KL penalty against the
unguided transition:
\begin{align}\label{eqn:masked-loss}
\mathcal{J}(\theta,\phi)
&=
\sum_{t=2}^{L}\sum_j
m_{t,j}(1-\widetilde{m}_{t,j})
\,\ell_\theta(x_{t,j};\widehat{\boldsymbol{z}}_t) +
\beta\sum_{t=2}^{L}
D_{\mathrm{KL}}\!\left(
q_\phi(\boldsymbol{z}_t\mid\mathbb{x},\mathbb{m})
\,\Vert\,
p_\theta(\boldsymbol{z}_t\mid
\widetilde{\boldsymbol{z}}_{t-1})
\right).
\end{align}
Here $\ell_\theta$ denotes the observation prediction loss.
Unlike the reconstruction term in \eqref{eqn:ELBO}, its prediction
is decoded from the corrected state $\widehat{\boldsymbol{z}}_t$
instead of an encoder sample $\widetilde{\boldsymbol{z}}_t$; as such, the resulting objective is a VAE-inspired surrogate
rather than an exact negative ELBO.

Through this masked training procedure, the model learns to
incorporate contemporaneous guidance when estimating the current
latent state under \textit{partial observability}. The mixed-frequency nowcasting problem described earlier is therefore a special case of this more general setup. More broadly, this mechanism naturally accommodates scenario generation, where a prescribed trajectory on a subset of coordinates acts as guidance, allowing the remaining variables to be coherently inferred.

\subsection{Irregularly spaced time series modeling and imputation}
\label{sec:irregular-ts}

Time series with irregularly spaced observations arise in various application domains including electronic health records, biology and ecology \citep{reinsel2003elements}. Traditionally, techniques such as linear/polynomial fitting, spline interpolation, time scale aggregation (e.g., binning, averaging) and SSM have been employed to handle such data \citep{reinsel2003elements,hamilton2020time}. More recently, neural network-based approaches have emerged, which can be broadly segmented into three categories: (1) preprocessing the input, by either mapping observations on a regularly spaced grid through discretization of the time interval \citep{marlin2012unsupervised,lipton2016directly}, or through interpolation \citep{shukla2019interpolation}; (2) using one of the regular neural network architectures (e.g., RNN or transformers) for time series modeling, with input augmented by information on missingness and intervals \citep{che2018recurrent}, or incorporating specialized modules such as time attention, which explicitly leverage temporal embeddings \citep{shukla2020multi}; and (3) modeling the latent space in continuous time, leveraging frameworks like neural ODEs \citep{rubanova2019latent,yildiz2019ode2vae}, which are naturally agnostic to the time scale. The remainder of this subsection focuses on representative approaches from the third category. 

The modeling framework proposed by \cite{rubanova2019latent}, detailed in Section~\ref{sec:neuralODE}, focuses on the application of neural ODE for irregularly spaced time series in tasks such as interpolation (i.e., imputation) and extrapolation (i.e., forecasting), respectively. At a high level, once the model is trained and yields estimated parameters $(\widehat{\theta}_f,\widehat{\theta}_g)$, the latent state trajectory can be computed by solving an initial value problem based on sampled initial value $\widetilde{\boldsymbol{z}}_{t_0}$ from its encoded distribution at desired time points:
\begin{equation*}
    \widehat{\boldsymbol{z}}_{t'_{1}}, \cdots, \widehat{\boldsymbol{z}}_{t'_{m}} = \text{ODESolve}\big(\widetilde{\boldsymbol{z}}_{t_0}, f_{\widehat{\theta}_f}, (t_0,t'_1,\cdots,t'_{m}) \big).
\end{equation*}
Here, $\{t'_1,\cdots,t'_{m}\}$ represent timestamps of interest at inference time---either imputation or prediction, distinct from the available ones $\{t_1,\cdots,t_n\}$ used for training the model. Once estimates of the latent states are available, the corresponding $\widehat{\boldsymbol{x}}_{t'_i}$'s can be readily obtained through the observation equation~\eqref{eqn:obs-ODE}. For both imputation and forecasting tasks, the training process follows the VAE-based pipeline outlined in Exhibit~\ref{odepipe}, with slight variations in how encoding and decoding are handled. Specifically,
\begin{itemize}[itemsep=0pt]
    \item For imputation, both encoding and the reconstruction error calculation are conducted based on the available time points $\{t_1,\cdots,t_n\}$;
    \item For forecasting, let $k=\floor*{n/2}$. The encoder traverses the observations at $t_1,\cdots,t_k$ backwards in time. The decoder solves the ODE forward at $t_{k+1},\cdots,t_n$, where the reconstruction error is calculated.
\end{itemize}
In a variation to~\cite{rubanova2019latent}, \cite{yildiz2019ode2vae} introduce a Bayesian neural second-order ODE to mitigate the incapability of a first-order ODE model to capture higher-order dynamics. Specifically, the latent state is modeled as:
\begin{equation*}
 \frac{\dd^2 \boldsymbol{z}_t}{\dd t^2} = h_{\theta_h}(\boldsymbol{z}_t, \frac{\dd \boldsymbol{z}_t}{\dd t}),
\end{equation*}
where $h_{\theta_h}: \mathbb{R}^d\times \mathbb{R}^d\mapsto \mathbb{R}^d$ is parameterized by a neural network. By considering the state as a tuple $\boldsymbol{z}_t:=(\boldsymbol{s}_t,\boldsymbol{v}_t)$ with $\boldsymbol{s}_t$ and $\boldsymbol{v}_t$ representing its position and velocity, the above equation can be equivalently written as a system of coupled first-order ODEs, namely,
\begin{align*}
    &\frac{\dd \boldsymbol{s}_t}{\dd t} = \boldsymbol{v}_t, \qquad \frac{\dd \boldsymbol{v}_t}{\dd t} = h_{\theta_h}(\boldsymbol{s}_t,\boldsymbol{v}_t),\\
    \Rightarrow~~~ &\boldsymbol{s}_t=\boldsymbol{s}_0 + \int_0^t \boldsymbol{v}_\tau \dd \tau, ~~\boldsymbol{v}_t = \boldsymbol{v}_0 + \int_0^t h_{\theta_h}(\boldsymbol{s}_\tau,\boldsymbol{v}_\tau)\dd \tau.
\end{align*}
The observation $\boldsymbol{x}_t$ depends only on the state position $\boldsymbol{s}_t$'s, and the velocity $\boldsymbol{v}_t$ serves as an auxiliary variable. Finally, model training, imputation and/or forecasting steps largely stay the same as the ones outlined for \cite{rubanova2019latent}. 

There is another line of work that does not leverage directly an SSM formulation, albeit the notion of latent process dynamics is still present. For example, \cite{de2019gru} utilize an RNN to implicitly model the latent state dynamics through GRU operations\footnote{Gated Recurrent Units (GRUs) are designed to model sequences efficiently by using update and reset gates to manage information flow. The update gate controls the amount of past information retained, while the reset gate the amount of past information ignored, allowing the model to manage long-term dependencies effectively.}, which when coupled with an ODE enables continuous-time modeling for irregularly spaced time series. In subsequent work, \cite{yuan2023ode} combine an SSM formulation with the ODE-RNN architecture in \cite{de2019gru} to address the same modeling task. Relatedly, \citet{kidger2020neural} introduce neural controlled differential equations (Neural CDEs), which construct a continuous control path through interpolation of irregularly sampled observations and use it to drive the evolution of a hidden state. This provides a continuous-time extension of recurrent modeling in which the hidden dynamics respond to the observed trajectory, without explicitly requiring a generative SSM formulation. Further details on these approaches can be found in the respective references.

\subsection{Scientific applications with mechanistically structured state-space models}

In many scientific applications, SSMs are used not only for prediction, but also to represent latent processes with mechanistic interpretation. In such settings, deep learning can play different roles. In some cases, neural components are introduced within an otherwise mechanistically specified model, for example to capture unknown forcing terms, nonlinear corrections, or partially observed transition mechanisms while preserving the core scientific structure. In other cases, the latent dynamics are modeled more flexibly with weaker structural assumptions, trading some mechanistic interpretability for greater representational capacity. This distinction is particularly relevant in scientific applications, where the latent states are often intended to retain causal or domain-specific meaning.

\paragraph{Epidemiology and infectious disease modeling.} A prominent example arises in compartmental epidemic models such as SIR or SEIR systems, where neural components can be embedded into otherwise mechanistic dynamics to account for unmodeled transmission effects, behavioral responses, or time-varying contact structure. In this setting, deep learning serves primarily to augment a scientifically motivated state-space specification rather than replace it entirely; see, for example, the universal differential equation framework of \citet{rackauckas2020universal}.

\paragraph{Ecology and environmental sciences.} Related ideas appear in ecological and environmental modeling, where latent-state formulations are often introduced to separate process variability from measurement noise. Here, neural parameterizations can be used to augment process-based models, for example by representing nonlinear interactions or unknown forcing terms, while retaining the broader scientific interpretation of the latent dynamics \citep{bonnaffe2021neural,madden2024deep,wesselkamp2024process}. Such applications are especially relevant because SSMs already play a central role in these domains (see, e.g., the review by \cite{auger2021guide}), and deep-learning-based extensions are only beginning to emerge.

\paragraph{Pharmacology and molecular kinetics.} In pharmacokinetic and pharmacodynamic applications, one often starts from mechanistic principles governing dose, concentration, and effect, and then uses neural components to increase flexibility where the classical specification is inadequate \citep{hussain2021neural,lu2021deep,joslyn2024pharmacologically}. Similar ideas also arise in molecular kinetics, where learned latent dynamical models can be used to approximate complex transition behavior while respecting key scientific structure \citep{mardt2018vampnets,chen2019nonlinear}.

Overall, these examples illustrate that, in scientific settings, the role of deep learning is often not to discard mechanistic state-space structure, but to extend it in a more flexible and data-adaptive way. This stands in contrast to settings where neural SSMs are used with weaker substantive assumptions and are primarily judged by predictive performance.

%% file: 06-conclusion.tex
\section{Concluding Remarks}
\label{sec:conclusion}


As noted throughout this review, classical approaches to estimating SSM parameters rely fundamentally on recursive filtering and smoothing calculations. To handle nonlinearities and/or non-Gaussianity in the model specification, one typically resorts either to deterministic approximations (e.g., extended Kalman and related Gaussian filters) or to simulation-based methods such as sequential Monte Carlo (SMC). Particle methods are appealing because they can accommodate highly flexible transition and observation mechanisms. While standard particle filters may deteriorate rapidly as the state dimension grows, more recent high-dimensional extensions, including block/local particle filters and their iterated variants for parameter estimation \citep{rebeschini2015can,li2024inference,ning2023iterated}, can mitigate this difficulty by introducing additional localization-type approximation structure.

Neural network-based approaches, by contrast, allow for highly flexible parameterizations of both the state and observation equations. Within this literature, VAE-type learning remains the dominant paradigm, typically combining an encoder and a decoder trained end-to-end. The encoder approximates the latent posterior through variational distributions such as $q_{\phi}(\boldsymbol{z}_t\mid \mathbb{x})$, often under simplifying assumptions such as mean-field or Gaussian structure, while the decoder specifies the corresponding generative model. Moreover, although VAE-based models are trained end-to-end, the underlying encoder and decoder architectures often still rely on recursive or sequential computation, for example through RNN-type modules.

It is worth noting that the discrete time linear SSM has also been extensively studied in the econometrics literature \citep{stock2012dynamic}, and is referred to as the exact Dynamic Factor Model, with the additional assumption of a diagonal covariance matrix of the error term in the observation equation. A focus in that stream of literature is to allow cross-sectional dependence in the error term of the observation equation, giving rise to the approximate Dynamic Factor Model (see review paper by \cite{stock2016dynamic} and references therein). Extensions addressing in addition temporal dependence in observation errors are provided in \cite{lin2020system}.

%% file: appendices.tex
\section{Discrete Time Linear Gaussian SSM}
\label{sec:discrete-linear-Gaussian-SSM}

This section first provides an overview of the key steps involved in estimating the model parameters based on the EM algorithm, wherein linearity and Gaussianity lead to closed form expressions for the E-step. Further, explicit expressions can be obtained for the M-step under the assumption that the model parameters are \textit{time invariant}. Finally, a variant of the SSM with regime-dependent coefficient matrices is discussed, highlighting the added challenges in estimation introduced by the dynamically evolving parameters.



A \textit{general} form of the model under consideration is given by
\begin{subequations}
\begin{align}
\text{state eqn:} & \quad \boldsymbol{z}_t  =  A_t \boldsymbol{z}_{t-1}+\boldsymbol{u}_t, \label{eq10.linear-ssm} \\
\text{obs. eqn:} & \quad \boldsymbol{x}_t  =  C_t \boldsymbol{z}_t+\boldsymbol{\epsilon}_t, \label{eq20.linear-ssm}
\end{align}
\end{subequations}
where $A_t\in\mathbb{R}^{d\times d}$ and $C_t\in\mathbb{R}^{p\times d}$ specify the autoregressive dynamics and the relationships between the latent state and the observed variables, respectively. The noise vectors are independent across time, with
\begin{equation*}
\begin{pmatrix}
\boldsymbol{u}_t \\
\boldsymbol{\epsilon}_t 
\end{pmatrix} 
\sim
\mathcal{N}\biggl(
\begin{pmatrix}
0 \\ 0 \end{pmatrix},
\begin{pmatrix}
Q_t & 0 \\ 0 & R_t
\end{pmatrix} 
\biggr).
\end{equation*}
Note that the equation shown in \eqref{eq10.linear-ssm} assumes the state to be a lag-1 vector autoregressive process. More lags can be accommodated in a straightforward manner, by rewriting a lag-r autoregressive process in its lag-1 form; see details in \cite{lutkepohl2005new}.


The computations in the E-step involve the following updates of (see also Section 2.1 of the main article): (i) the \textit{filtering step}, which calculates the posterior distribution of the latent state given the observed trajectory up to time $t$---$p_{{\theta}}(\boldsymbol{z}_t | \mathbb{x}_{1:t})$ in equation~\eqref{eq:filtering-equation}, (ii) the \textit{prediction step}, which calculates the predictive distribution of the next latent state given the observed trajectory up to time $t-1$---$p_{{\theta}}(\boldsymbol{z}_t | \mathbb{x}_{1:t-1})$, and (iii) the \textit{smoothing step} that updates the posterior distribution of the latent state given all the data---$p_{{\theta}}(\boldsymbol{z}_t | \mathbb{x}_{1:T})$ in equation~\eqref{eq:backward-smoothing-recursion}.

Let $\boldsymbol{z}_{t|s}:= \mathbb{E}(\boldsymbol{z}_t|\mathbb{x}_{1:s})$ and $P_{t|s}:=\text{Cov}(\boldsymbol{z}_t|\mathbb{x}_{1:s}), \ s, t=0, 1, 2, \cdots, T$. For the linear Gaussian SSM, all three distributions computed during the E-step are \textit{Gaussian}, whose means and covariance matrices can be sequentially updated based on the following recursive formulas, for \textit{given} values of the model parameters $\theta_t=(A_t, C_t, Q_t, R_t)$. Concretely,
\begin{itemize}[itemsep=0pt]
\item Filtering step:
\begin{align*}
\boldsymbol{z}_{t|t} & = \boldsymbol{z}_{t|t-1}+K_t(\boldsymbol{x}_t-C_t\boldsymbol{z}_{t|t-1}), \\ 
P_{t|t} & =  (I- K_t C_t) P_{t|t-1},
\end{align*}
where $K_t$ corresponds to the \textit{Kalman gain} matrix given by
\begin{equation*}
K_t=P_{t|t-1}C^\top_t (C_tP_{t|t-1} C^\top_t + R_t)^{-1}.
\end{equation*}
\item Prediction step:
\begin{align*}
\boldsymbol{z}_{t|t-1} & = A_t\boldsymbol{z}_{t-1|t-1},\\
P_{t|t-1} & = A_t P_{t-1|t-1}A_t^\top+Q_t.
\end{align*}
\item Smoothing step: 
\begin{align*}
\boldsymbol{z}_{t|T} & = \boldsymbol{z}_{t|t}+H_t (\boldsymbol{z}_{t+1|T}-\boldsymbol{z}_{t+1|t}),\\
P_{t|T} & = P_{t|t}+H_t(P_{t+1|T}-P_{t+1|t})H_t^\top,
\end{align*}
with $H_t=P_{t|t}A_{t+1}^\top P^{-1}_{t+1|t}$ being the Kalman smoother gain.
\end{itemize}
Derivations of these formulas, collectively known as the Kalman filter algorithm, can be found in standard references \citep[e.g., in][]{harvey1990forecasting,durbin2012time}.

While the Kalman filter algorithm solves the state identification problem for the general form of the linear Gaussian SSM, estimation of the parameters $\theta_t$ is intractable without additional simplification in the model specification, due to the fact that a single observation is available for each value of the time varying parameter. To overcome this limitation, the most widely used version of the model assumes \textit{time invariant} parameters; i.e., $\theta_t\equiv\theta=(A, C, Q, R)$. In that case, the M-step computations yield closed form expressions for the model parameters $\theta$, due to the Gaussian nature of the expected log-likelihood $L(\theta,\theta^{(k)})$ (the $\mathcal{Q}(\theta\mid \theta^{(k)})$ surrogate objective in the main manuscript). Specifically, the M-step estimates are then given by \citep{shumway1982approach}: 
{\small
\begin{align*}
\widehat{A} &=\bigl(\sum\nolimits_{t=1}^T (V_{t,t-1|T} + \boldsymbol{z}_{t|T}\boldsymbol{z}^\top_{t-1|T})\bigr)\big(\sum\nolimits_{t=1}^T (P_{t-1|T}+\boldsymbol{z}_{t-1|T}\boldsymbol{z}^\top_{t-1|T})\big)^{-1}\\
\widehat{C} &=\bigl(\sum\nolimits_{t=1}^T \boldsymbol{x}_t \boldsymbol{z}^\top_{t|T}\bigr)\bigl(\sum\nolimits_{t=1}^T (P_{t|T}+\boldsymbol{z}_{t|T}\boldsymbol{z}^\top_{t|T})\bigr)^{-1}, \\
\widehat{Q} &= \frac{1}{T}\sum\nolimits_{t=1}^T
\Bigl[P_{t|T}+\boldsymbol{z}_{t|T}\boldsymbol{z}_{t|T}^{\top}-\widehat{A}\bigl(V_{t,t-1|T}^{\top}+\boldsymbol{z}_{t-1|T}\boldsymbol{z}_{t|T}^{\top}\bigr)\Bigr],\\
\widehat{R}&=\frac{1}{T}\sum\nolimits_{t=1}^T \bigl(\boldsymbol{x}_{t}\boldsymbol{x}^\top_{t} - \widehat{C} \boldsymbol{z}_{t|T}\boldsymbol{x}_t^\top\bigr),
\end{align*}
}%
where $V_{t,t-1|T}:=\text{Cov}(\boldsymbol{z}_t,\boldsymbol{z}_{t-1}|\mathbb{x}_{1:T})$ and can be computed recursively based on the updates of the quantities from the Kalman filter as follows, $t=1,\cdots,T-1$:
\begin{equation*}
V_{t,t-1|T}=P_{t|t}J^\top_{t-1}+J_{t}(V_{t+1,t|T}-AP_{t|t})J^\top_{t-1},
\end{equation*}
with $J_{t}=P_{t|t}A^\top (P_{t+1|t})^{-1}$ and 
\begin{equation*}
    V_{T,T-1|T}=(I-K_TC)AP_{T-1|T-1}.
\end{equation*}
Finally, note that another estimation strategy for $\theta$ based on the likelihood marginalization approach is given in Chapter 6 of~\cite{shumway2000time}.

\paragraph{Linear Gaussian switching SSM.\ } This variant exhibits richer dynamics compared to the time-invariant model, while parameter estimation remains feasible, albeit significantly more technically involved. The model includes an \textit{additional latent state} variable $S_t$ that takes values in a finite state space $S_t\in \mathcal{S}=\{s_1,\cdots,s_M\}$ and follows Markovian dynamics governed by a transition kernel $\mathcal{P}_S:=\mathbb{P}(S_t|S_{t-1})$. Each state $s_m\in\mathcal{S}$ has its associated transition and covariance matrices $(A_{s_m}, C_{s_m}, Q_{s_m}, R_{s_m})$. The state and observation equations of this model variant are given by
\begin{align*}
\boldsymbol{z}_t &=  A_{S_t} \boldsymbol{z}_{t-1}+\boldsymbol{u}_{S_t},   \\
\boldsymbol{x}_t &=  C_{S_t} \boldsymbol{z}_t+\boldsymbol{\epsilon}_{S_t}. 
\end{align*}
This model combines the linear dynamics of Gaussian SSMs, with a discrete structure of Hidden Markov Models for the time varying parameters.

The model has been studied in the systems engineering \citep{chang1978state}, econometrics \citep{hamilton1989new} and statistics \citep{shumway1991dynamic} literature. Parameter learning is based on the EM algorithm. Note that in the filtering step, the posterior distribution of the \textit{augmented} latent state $(\boldsymbol{z}_t, S_t)$ given an observed trajectory needs to be computed. As shown in \cite{chang1978state}, even with a \textit{known} transition kernel $\mathcal{P}_S$, this posterior distribution corresponds to a Gaussian mixture distribution with $M^T$ components, where $T$ is the length of the observed trajectory. As such, direct computation is intractable. In addition, predicting $\boldsymbol{z}_t$ not only depends on the extra state variable $S_t$, but also on its value at time $t-1$, and this further increases the computational burden; i.e.,
$\boldsymbol{z}_{t|t}(m,m'):=\mathbb{E}(\boldsymbol{z}_t|\mathbb{x}_{1:t}, S_t=m, S_{t-1}=m'), \ m, m'\in\mathcal{S},$ and analogously for $P_{t|t}(m,m'):=\text{Cov}(\boldsymbol{z}_t|\mathbb{x}_{1:t}, S_t=m, S_{t-1}=m')$. To render the computations in the E-step manageable, \cite{kim1994dynamic} and \cite{bar2004estimation} introduce appropriate approximations, originally proposed in \cite{harrison1976bayesian}, enabling computations whose 
complexity is linear in $M$ rather than $M^T$. They also derive the corresponding filtering, prediction and smoothing updates for state identification and maximum likelihood estimators for the model parameters. Alternatively, \cite{ghahramani2000variational} leverage a variational approach, using a tractable distribution that eliminates some of the dependencies between the original state $\boldsymbol{z}_t$ and the extra state $S_t$ to approximate the posterior distribution in the filtering step. A number of Bayesian approaches have also been proposed for estimating this model's parameters, including \cite{carter1994gibbs, carter1996markov,kim1998business,kim1999state,ghahramani2000variational,fruhwirth2001fully,petetin2021structured}.

\paragraph{The discrete process SSM.} A more involved variant of the discrete time linear Gaussian SSM, known as the discrete process model
\citep{roesser1975discrete,kitagawa2012smoothness}, is defined as
\begin{align*}
\boldsymbol{z}_t &= A\boldsymbol{z}_{t-1}+ \boldsymbol{u}_t,\\
\boldsymbol{x}_t &\sim p_{\theta}(\boldsymbol{x}_t|\boldsymbol{z}_t),  \ \ \boldsymbol{x}_t \in \mathbb{Z}^p,
\end{align*}
where the state equation is linear, and the observation process takes discrete values, so that  it can handle binary or count data. Parameter learning is based on  the EM algorithm. However, the presence of a nonlinear, non-Gaussian observation equation introduces complications, necessitating the use of the general learning framework discussed in Section 2.2, instead of the straightforward updates offered by the Kalman filter; see \cite{kitagawa1987non,kitagawa1988numerical} for further details.

\section{The Extended Kalman, the Gaussian and the Unscented Kalman filters}
\label{sec:EKF}

This section outlines three key filtering techniques---Extended Kalman Filter (EKF), Gaussian Filter, and Unscented Kalman Filter (UKF)---commonly used for nonlinear state-space models. It highlights their respective methodologies, and briefly discusses their strengths, and limitations in approximating the posterior distributions required in state identification.

We consider a discrete time non-linear SSM with \textit{additive Gaussian} state and observation noise, given by 
\begin{subequations}
\begin{align}
\text{initial state:}\quad & \boldsymbol{z}_0 = f(\boldsymbol{u}_0; \theta); \label{eq:nl-initial-state} \\
\text{state eqn:}\quad &\boldsymbol{z}_t = f(\boldsymbol{z}_{t-1}; \theta)+\boldsymbol{u}_t,  \label{eq:nl-state-eqn} \\
    \text{obs. eqn:}\quad &\boldsymbol{x}_t = g(\boldsymbol{z}_t; \theta)+\boldsymbol{\epsilon}_t,  \label{eq:nl-obs-eqn}
\end{align}
\end{subequations}
where $\{\boldsymbol{u}_t\}, \{\boldsymbol{\epsilon}_t\}$ are sequences of Gaussian independent and identically distributed random variables of shocks/noises, with 
\[\begin{pmatrix}
\boldsymbol{u}_t \\
\boldsymbol{\epsilon}_t 
\end{pmatrix} 
\overset{iid}{\sim}
\mathcal{N}\biggl(
\begin{pmatrix}
0 \\ 0 \end{pmatrix},
\begin{pmatrix}
Q & 0 \\ 0 & R
\end{pmatrix} 
\biggr).
\]
For ease of presentation, we further assume that $f, g$ are deterministic \textit{time invariant} multivariate functions of appropriate dimensions.

\paragraph{Extended Kalman filter.} The key idea underlying this filter is to assume that the posterior distributions appearing in the filtering, prediction and smoothing steps are \textit{approximately Gaussian}, whose means and covariance matrices are obtained by \textit{linearizing} the state and observation functions using first-order Taylor expansions. 

Let $J_f(\cdot):=\frac{\partial f}{\partial \boldsymbol{z}}$ and $J_g(\cdot):=\frac{\partial g}{\partial \boldsymbol{z}}$ be the Jacobians. The recursive formulas for the filtering, prediction and smoothing steps are then given by:
\begin{itemize}[itemsep=0pt]
\item Filtering step:
\begin{align*}
\boldsymbol{z}_{t|t} & =  \boldsymbol{z}_{t|t-1}+K_t\big(\boldsymbol{x}_t-g(\boldsymbol{z}_{t|t-1})\big), \\ \nonumber
P_{t|t} & =  \bigl(I - K_t J_g(\boldsymbol{z}_{t|t-1})\bigr) P_{t|t-1},
\end{align*}
where 
{\small
\begin{equation*}
    K_t=P_{t|t-1}J^\top_g(\boldsymbol{z}_{t|t-1}) \big[J_g(\boldsymbol{z}_{t|t-1}) P_{t|t-1} J_g^\top(\boldsymbol{z}_{t|t-1}) + R\big]^{-1}.
\end{equation*}
}%
\item Prediction step:
\begin{align*}
\boldsymbol{z}_{t|t-1} & = f(\boldsymbol{z}_{t-1|t-1}), \\
P_{t|t-1} & = J_f(\boldsymbol{z}_{t-1|t-1}) P_{t-1|t-1} 
J^\top_f(\boldsymbol{z}_{t-1|t-1})+Q.
\end{align*}
\item Smoothing step: 
\begin{align*}
\boldsymbol{z}_{t|T} & = \boldsymbol{z}_{t|t}+H_t (\boldsymbol{z}_{t+1|T}-\boldsymbol{z}_{t+1|t}), \\
P_{t|T} & = P_{t|t}+H_t(P_{t+1|T}-P_{t+1|t})H_t^\top,
\end{align*}
with $H_t=P_{t|t}J^\top_f(\boldsymbol{z}_{t|t}) P^{-1}_{t+1|t}$.
\end{itemize}
It can be seen that the filtering, smoothing and prediction updates of the EKF for the posited Gaussian additive model, follow closely those for the Kalman filter for the linear Gaussian model. Specifically, the coefficient matrices $A$ and $C$ are replaced by the functions $f$ and $g$, respectively, in the updates of the mean $\boldsymbol{z}_{t|s}$, and by the Jacobian matrices $J_f$ and $J_g$ of the $f$ and $g$ functions evaluated at the corresponding expectation $\boldsymbol{z}_{t|s}$, for the updates of the covariances $P_{t|s}$.  

\begin{remark} 
The EKF updates presented above use first-order Taylor expansions of the functions $f$ and $g$. Second-order Taylor expansions have also been considered in the literature \citep{gelb1974applied,roth2011efficient}, although they become computationally expensive for larger size SSMs. Further, iterative EKF schemes \citep{gelb1974applied,bell1993iterated} involve an additional pass after an initial set of latent states estimates, and use the same EKF updates to enhance accuracy.
\end{remark}

\begin{remark}
In the case where the error/noise terms do not enter the model in an additive form, analogous EKF updates can be derived for the general nonlinear SSM given in \eqref{eq35.initial-state}-\eqref{eq50:obs-eqn}, with Gaussian state and observation error terms. At the high level, key modifications include: (i) in expressions involving the functions $f(\boldsymbol{z},\boldsymbol{u})$ and $g(\boldsymbol{z},\boldsymbol{\epsilon})$, the noise variables should be set to zero; (ii) the Jacobian matrices $J_f$ and $J_g$ incorporate the error terms and when evaluated, a value of zero is also used for the corresponding error terms.
\end{remark}

The main advantage of the EKF over more general SMC methods is its relative simplicity. Linearization is a standard way of constructing approximations to non-linear systems, making EKF easy to apply. Further, when the degree of non-linearity of the SSM is mild, EKF performs well, especially with advanced iterative schemes. However, if the system exhibits strong nonlinearities in the state and observation equations, EKF performance may degrade significantly.

\paragraph{Gaussian filters.} To accommodate \textit{strong} non-linearities, this family of filters was proposed in \cite{ito2000gaussian,wu2006numerical}. The starting point is once again that the posterior distributions required in the E-step are assumed to be \textit{approximately Gaussian}. Hence, the means $\boldsymbol{z}_{t|t}$ and the covariance matrices $P_{t|t}$ of the approximating Gaussian distributions need to be computed as shown next.
\begin{itemize}[itemsep=0pt]
\item Prediction step. The key assumption made is that the posterior distribution of the random variable $\boldsymbol{z}_{t-1}|\mathbb{x}_{1:t-1}$  is Gaussian with mean $\boldsymbol{z}_{t-1|t-1}$ and covariance matrix $P_{t-1|t-1}$, i.e., $p_{\theta}(\boldsymbol{z}_{t-1}|\mathbb{x}_{1:t-1})=\mathcal{N}(\boldsymbol{z}_{t-1|t-1},P_{t-1|t-1})$. Using the state equation $\boldsymbol{z}_t=f(\boldsymbol{z}_{t-1})+\boldsymbol{u}_t$, we compute the first two moments of the predicted state distribution $\boldsymbol{z}_t|\mathbb{x}_{1:t-1}$ as follows:
{\small
\begin{align*}
\boldsymbol{z}_{t|t-1} & = \mathbb{E}(\boldsymbol{z}_t|\mathbb{x}_{1:t-1}) =  \int_{\mathbb{R}^d} f(\boldsymbol{z}_{t-1}) p_{\theta}(\boldsymbol{z}_{t-1}|\mathbb{x}_{1:t-1})\dd\boldsymbol{z}_{t-1}, \\ 
P_{t|t-1} & = \text{Cov} (\boldsymbol{z}_{t}|\mathbb{x}_{1:t-1}) = Q+
\int_{\mathbb{R}^d} (f(\boldsymbol{z}_{t-1})-\boldsymbol{z}_{t|t-1})(f(\boldsymbol{z}_{t-1})-\boldsymbol{z}_{t|t-1})^\top p_{\theta}(\boldsymbol{z}_{t-1}|\mathbb{x}_{1:t-1})\dd\boldsymbol{z}_{t-1}.
\end{align*}
}%
\item Filtering step. The starting point is to use the \textit{Gaussian approximation} of the predicted density that  we obtained from the prediction step; namely, $p_{\theta}(\boldsymbol{z}_{t}|\mathbb{x}_{1:t-1})=\mathcal{N}(\boldsymbol{z}_{t|t-1},P_{t|t-1})$. Then, combined with the observation equation $\boldsymbol{x}_t=g(\boldsymbol{z}_t)+\boldsymbol{\epsilon}_t$, we can obtain a Gaussian approximation of the filtering distribution of $\boldsymbol{z}_t|\mathbb{x}_{1:t}$. However, this is computationally challenging \citep{ito2000gaussian}. Instead, an efficient moment-matching based strategy approximates the \textit{joint} conditional distribution of the latent state and the observation $[\boldsymbol{z}_t, \boldsymbol{x}_t]^\top | \mathbb{x}_{1:t-1}$ as Gaussian. Subsequently, using a conditioning argument together with the properties of the Gaussian distribution, we can obtain the first two moments of the quantity of interest, namely $\boldsymbol{z}_t|[\boldsymbol{x}_t, \mathbb{x}_{1:t-1}]=\boldsymbol{z}_t|\mathbb{x}_{1:t}$. Consider the joint conditional distribution of $[\boldsymbol{z}_t, \boldsymbol{x}_t]^\top | \mathbb{x}_{1:t-1}$
\begin{equation*}
\begin{pmatrix}
    \boldsymbol{z}_t \\ \boldsymbol{x}_t
\end{pmatrix} \Big|\mathbb{x}_{1:t-1} \sim \mathcal{N}
\biggl(\begin{pmatrix}
    \boldsymbol{z}_{t|t-1} \\ \widehat{\boldsymbol{x}}_{t|t-1} 
\end{pmatrix},
\begin{pmatrix}
P_{t|t-1} & \Sigma_{zx,t} \\ \Sigma_{xz,t} & \Sigma_{xx,t}    
\end{pmatrix}\biggr).
\end{equation*}
It can be seen that the following quantities need to be  computed: (i) the expected value of the observation measurement, denoted by $\widehat{\boldsymbol{x}}_{t|t-1}$; (ii) its covariance matrix, denoted by $\Sigma_{xx,t}$ of $\boldsymbol{x}_t$, and (iii) the cross-covariance of $\boldsymbol{x}_t$ with $\boldsymbol{z}_t$, denoted by $\Sigma_{zx,t}$. Note that the expected value of $\boldsymbol{z}_t|\mathbb{x}_{1:t-1}$ has already been obtained from the prediction step. We then have
{\small
\begin{align*}
\widehat{\boldsymbol{x}}_{t|t-1} & = \int_{\mathbb{R}^d} g(\boldsymbol{z}_{t}) p_{\theta}(\boldsymbol{z}_{t}|\mathbb{x}_{1:t-1})\dd\boldsymbol{z}_{t}, \\ 
\Sigma_{xx,t} & =  R +
\int_{\mathbb{R}^d} (g(\boldsymbol{z}_{t})-\widehat{\boldsymbol{x}}_{t|t-1})(g(\boldsymbol{z}_{t})-\widehat{\boldsymbol{x}}_{t|t-1})^\top p_{\theta}(\boldsymbol{z}_{t}|\mathbb{x}_{1:t-1})\dd\boldsymbol{z}_{t}, \\ 
\Sigma_{zx,t} & =   
\int_{\mathbb{R}^d} (\boldsymbol{z}_{t}-\boldsymbol{z}_{t|t-1})(g(\boldsymbol{z}_{t})-\widehat{\boldsymbol{x}}_{t|t-1})^\top p_{\theta}(\boldsymbol{z}_{t}|\mathbb{x}_{1:t-1})\dd\boldsymbol{z}_{t}.
\end{align*}
}%
Finally, based on these quantities and the formulas for
the conditional expected value and covariance of a multivariate Gaussian distribution, we get the filtering updates:
\begin{align*}
\boldsymbol{z}_{t|t} & =  \boldsymbol{z}_{t|t-1}+K_t(\boldsymbol{x}_t-\widehat{\boldsymbol{x}}_{t|t-1}), \\
P_{t|t} & =  P_{t|t-1} -  K_t\Sigma_{xx,t} K^\top_t,
\end{align*}
where $K_t=\Sigma_{zx,t}\Sigma^{-1}_{xx,t}$.
\item 
Smoothing step: Analogous derivations to those in the filtering step lead to the following recursive updates \citep{sarkka2010gaussian}
\begin{align*}
\boldsymbol{z}_{t|T} & =  \boldsymbol{z}_{t|t}+H_t(\boldsymbol{z}_{t+1|T}-\boldsymbol{z}_{t+1|t}), \\ 
P_{t|T} & =  P_{t|t} + H_t(P_{t+1|T} - P_{t+1|t}) H_t^\top,
\end{align*}
where $H_t= L_{t,t+1} P^{-1}_{t+1|t}$, and $L_{t,t+1}=\text{Cov}\bigl(\boldsymbol{z}_t-\boldsymbol{z}_{t|t}, f(\boldsymbol{z}_t)-\boldsymbol{z}_{t+1|t} \bigr)$; the latter is computed based on the Gaussian filtering distribution with mean $\boldsymbol{z}_{t|t}$ and covariance $P_{t|t}$.
\end{itemize}
A key difference compared to the EKF is that in the filtering step, instead of $g(\boldsymbol{z}_{t|t-1})$ (that appears on the RHS of the filtering update), an \textit{approximation} based on $\widehat{\boldsymbol{x}}_{t|t-1}$ is used, and analogously for the quantities that appear in the Kalman gain matrix $K_t$.

Implementing the Gaussian filter involves calculating five integrals, two during the prediction step and three during the filtering step. These integrals can be computed efficiently with established numerical integration methods like Gauss-Hermite quadrature \citep{ito2000gaussian} or cubature \citep{wu2006numerical}, and/or Monte Carlo techniques.

\paragraph{Unscented Kalman filter.} It is based on the Gaussian filter’s moment-matching approach, with a key distinction lying in how the integrals in the prediction and filtering steps are computed. The main idea is to compute the required integrals using a weighted sum of carefully chosen ``sigma" points based on the unscented transform \citep{julier2004unscented}, designed to match the mean and covariance of the distribution at time $t-1$. Note that unlike Monte Carlo methods, these sigma points are selected deterministically. In the prediction step, the selected sigma points are transformed based on the function $f$, based on which the mean and covariance of the updated posterior distribution are computed. Analogous steps are used for computing the filtering update \citep[see][for more details]{julier2004unscented}.

Unlike the EKF, the Gaussian filter and its UKF variant do not require differentiability of the $f$ and $g$ functions, and avoid computing Jacobian matrices. However, both can be computationally more expensive than the EKF, especially for larger scale SSMs. However, empirical evidence suggests that they typically yield better performance, especially for strongly nonlinear systems.


\section{Nonlinear SDE based SSMs}\label{sec:ct-SSM}

This section provides a brief review regarding SSMs defined by nonlinear SDEs. Given the complexity of such a model, we note that under the classical framework, the system identification problem---that encompasses both state identification and parameter estimation---is computationally challenging. On the other hand,  the following two problems: (i) the state identification problem for \textit{fixed and known} model parameters, and (ii) the parameter estimation problem for an observed SDE in the form of $\dd\boldsymbol{x}_t=g(\boldsymbol{x}_t)\dd t+\dd\boldsymbol{w}^x(t)$ (note the absence of a latent state) have been studied in their respective context. Arguably, at the conceptual level, one can perform estimation akin to an EM algorithm, by alternating between the state identification and parameter estimation, with the latter considering an augmented system wherein the latent variables are imputed by the former.

To this end, the remainder of the section provides a brief review of the state identification problem, emphasizing the technical requirements that filters satisfy and also presenting the Kalman filter updates for linear SDE based SSMs. Additionally, it briefly discusses and references the limited literature on the parameter estimation for observed nonlinear SDE-based models. 

\paragraph{The state identification problem.} Let $(\Omega,\mathcal{F},\{\mathcal{F}_t\}_{t\in\mathbb{R}_+},\mathbb{P})$ denote a filtered probability space and consider a $(d+p)$ dimensional process $[\boldsymbol{z}_t, \boldsymbol{x}_t]^\top$, adapted to the filtration $\mathcal{F}_t$. The SSM is defined by the following two stochastic differential equations:
\begin{subequations}
\begin{align}
\text{state eqn:}~&\dd\boldsymbol{z}(t) = f(\boldsymbol{z}(t))\dd t+\sigma(\boldsymbol{z}(t))\dd \boldsymbol{w}^{z}(t), 
\label{eq.10:state-ct-SSM} \\
\text{obs eqn:}~&\dd\boldsymbol{x}(t) = g(\boldsymbol{z}(t))\dd t+\dd \boldsymbol{w}^{x}(t),  \label{eq.20:obs-ct-SSM}
\end{align}
\end{subequations}
where $f(\cdot): \mathbb{R}^d\rightarrow\mathbb{R}^d$, 
$\sigma(\cdot): \mathbb{R}^d\rightarrow\mathbb{R}^{d\times d}$,
and $g(\cdot): \mathbb{R}^d\rightarrow\mathbb{R}^p$ are $\mathcal{F}$-measurable functions and $\boldsymbol{w}^z, \boldsymbol{w}^x$ independent 
$\mathbb{R}^d$ and $\mathbb{R}^p$-valued standard Wiener processes, respectively. In addition, the initial value of the latent state $\boldsymbol{z}_0$ is a random variable drawn from a distribution which is independent of the Wiener processes $\boldsymbol{w}^z$ and $\boldsymbol{w}^x$. Further, it is assumed that $f, \sigma$ are globally Lipschitz functions and $g$ satisfies the linear growth condition $\|g(z)\|^2_2\leq c(1+\|z\|^2_2)$ for some constant $c>0$. The former condition ensures pathwise uniqueness of the solution, while the latter ensures that the solution of \eqref{eq.20:obs-ct-SSM} does not explode in finite time almost surely \citep{ikeda2014stochastic}.

The state estimation (filtering) problem (for given $f, \sigma, g$) is then defined by determining the probability distribution of the state $\boldsymbol{z}(t)$ conditional on the filtration generated by observations $\mathbb{x}_{0:t}:=\{\boldsymbol{x}(\tau), \tau\in [0,t]\}$. Concretely, the goal is to calculate the probability kernel
\begin{equation*}
\pi_{t|s}(\dd z,\mathbb{x}_{0:s}):=\mathbb{P}(\boldsymbol{z}(t)\in \dd z \ | \ \mathbb{x}_{0:s}).
\end{equation*}
Recall from Section \ref{sec:SSM-learning} that the state estimation problem comprises the following three tasks: learning  $\pi_{t|s}(\dd z,\mathbb{x}_{0:s})$ for (i) $s=t$ (\textit{filtering} problem), (ii) $s>t$ (\textit{smoothing} problem) and (iii) $s<t$ (\textit{prediction} problem) \citep{bain2009fundamentals}. For the filtering problem, the kernel $\pi_{t|t}$ satisfies the Kushner-Stratonovich stochastic differential equation (SDE).

Specifically, for a bounded twice differentiable function $h: \mathbb{R}^d\rightarrow\mathbb{R}$, let $\mathfrak{V}_t h$ denote the infinitesimal generator of the process $\boldsymbol{z}$, given by
\begin{equation*}
\mathfrak{V}_t h:=\sum_{i=1}^d f_i(\boldsymbol{z}) \frac{\partial h(\boldsymbol{z})}{\partial{\boldsymbol{z}_i}} + \frac{1}{2} \sum_{i,j=1}^d [\sigma(\boldsymbol{z})\sigma(\boldsymbol{z})^\top]_{ij} \frac{\partial^2 h(\boldsymbol{z})}{\partial{\boldsymbol{z}_i}
\partial{\boldsymbol{z}_j}}.
\end{equation*}
The Kushner-Stratonovich \textit{nonlinear} SDE for $h$ is then given by
\begin{align}\label{eq.40:KS-sde}
\dd \pi_{t|t}(h,\mathbb{x}_{0:t})
= \pi_{t|t}(\mathfrak{V}_t h,\mathbb{x}_{0:t})\,\dd t + \bigl[
    \pi_{t|t}(hg,\mathbb{x}_{0:t})
    - \pi_{t|t}(h,\mathbb{x}_{0:t})\,
      \pi_{t|t}(g,\mathbb{x}_{0:t})
\bigr]^{\top}
\bigl(
    \dd \boldsymbol{x}(t)
    - \pi_{t|t}(g,\mathbb{x}_{0:t})\,\dd t
\bigr).
\end{align}
Here $g:\mathbb{R}^d\to\mathbb{R}^p$, and $hg$ denotes the vector-valued function whose $j$th component is $h(z)g_j(z)$. The expression in the last parenthesis represents the \emph{innovation increment}, namely the difference between the observed increment $\dd \boldsymbol{x}(t)$ and its conditional expectation $\pi_{t|t}(g)\,\dd t$, given the observation history. The Kushner-Stratonovich SDE is a nonlinear one, which in general poses significant challenges in solving it. Let $\mathbb Q$ be the reference measure under which
$\boldsymbol{x}(t)$ is a standard Wiener process, and define
\[
\Lambda_t =
\frac{\mathrm{d}(\mathbb P|_{\mathcal F_t})}
     {\mathrm{d}(\mathbb Q|_{\mathcal F_t})}.
\]
The unnormalized filter is
\[
q_{t|t}(h,\mathbb{x}_{0:t})
:= \mathbb E_{\mathbb Q}\!\left[
    \Lambda_t h(\boldsymbol{z}(t)) \mid \mathbb{x}_{0:t}
\right].
\]
By the Kallianpur--Striebel formula,
\[
\pi_{t|t}(h,\mathbb{x}_{0:t})
= \frac{q_{t|t}(h,\mathbb{x}_{0:t})}
       {q_{t|t}(1,\mathbb{x}_{0:t})},
\]
where $1$ denotes the constant function $h(z)=1$. Then, the Zakai SDE for the function $h$ is given by 
\begin{equation*}
\dd q_{t|t}(h,\mathbb{x}_{0:t})=q_{t|t}(\mathfrak{V}_t h, \mathbb{x}_{0:t})\dd t+q_{t|t}(hg, \mathbb{x}_{0:t})^\top\dd\boldsymbol{x}_t.
\end{equation*}
The counterparts of the Kushner-Stratonovich and Zakai SDEs for the prediction and the smoothing tasks are more involved and provided in \cite{pardoux1981non,steinberg1994fixed}. Further, the smoothing task is more technically involved, since it requires the introduction of backward stochastic integration.

The characterization of the solutions and their uniqueness of the Kushner-Stratonovich and Zakai equations has been the topic of theoretical investigation \citep{rozovskii1991simple,bain2009fundamentals}, since it is of central importance for their solution by numerical methods. Note that in general $\pi_{t|t}(\cdot,\mathbb{x}_{0:t})$ (the solution to \eqref{eq.40:KS-sde}) is an \textit{infinite dimensional} process \citep{legland2005splitting,ocone2006lie}. The most notable exception is the Kalman-Bucy filter \citep{kalman1961new} used for the state identification problem of an SSM whose specification is given by a \textit{linear SDE}.

\paragraph{SSMs based on linear SDEs and the Kalman-Bucy filter.} In this case, \eqref{eq.10:state-ct-SSM}-\eqref{eq.20:obs-ct-SSM} become
\begin{subequations}
\begin{align}
\text{state eqn:} &\quad \dd\boldsymbol{z}(t) = A\boldsymbol{z}(t) \dd t + Q \dd \boldsymbol{w}^{z}(t)  \label{eq.60:state-linear-ct-SSM} \\
\text{obs. eqn:}&\quad \dd\boldsymbol{x}(t) =C\boldsymbol{z}(t) \dd t + \dd \boldsymbol{w}^{x}(t),  \label{eq.70:obs-ct-SSM}
\end{align}
\end{subequations}
with matrix coefficients $A, Q\in\mathbb{R}^{d\times d}$, and $C\in\mathbb{R}^{p\times d}$.

With a Gaussian initial state, the conditional distribution $\pi_{t|s}(\cdot,\mathbb{x}_{0:s})$ of the latent state $\boldsymbol{z}_t$ given an observed trajectory, corresponds to a multivariate (finite dimensional) Gaussian distribution, characterized by its first and second moments alone.
Specifically, for coordinate $j=1,\cdots,d$ of the state $\boldsymbol{z}(t)$, define its conditional mean as $\boldsymbol{z}^j_{t|s}:=\mathbb{E}(\boldsymbol{z}^j(t)|\mathbb{x}_{0:s})$
and its conditional covariance matrix by 
\begin{equation*}
    P^{jk}_{t|s}:=\mathbb{E}(\boldsymbol{z}^j(t)\boldsymbol{z}^k(t)|\mathbb{x}_{0:s})-\boldsymbol{z}^j_{t|s}\boldsymbol{z}^k_{t|s}.
\end{equation*}
The $d$-dimensional vector $\boldsymbol{z}_{t|t}$ then satisfies the following SDE
\begin{equation*}
\dd\boldsymbol{z}_{t|t}=A\boldsymbol{z}_{t|t} \dd t + K_t \bigl(\dd\boldsymbol{x}(t)-
C\boldsymbol{z}_{t|t}\dd t\bigr),
\end{equation*}
and the $d\times d$ matrix $P_{t|t}$ satisfies the \textit{deterministic} matrix Riccati equation (ODE)
\begin{align*}
\dd P_{t|t} &=QQ^\top\dd t +A P_{t|t} \dd t+P_{t|t} A^\top \dd t- P_{t|t} C^\top CP_{t|t} \dd t,
\end{align*}
with $K_t=P_{t|t}C^\top$ being the Kalman gain matrix. Of note, the SDE for the conditional mean and the ODE for the conditional variance are also satisfied for linear SSMs with time-varying coefficient matrices $A_t,C_t,Q_t$ and identity observation-noise covariance (recall that these parameters are assumed to be known in the context of state identification).

\paragraph{Model parameter estimation for observed diffusion processes.} 

The starting point is that, for sufficiently small time increments $\Delta t$ and under certain regularity conditions \citep{kloeden1992stochastic}, the distribution of the increment
$\boldsymbol{x}(t+\Delta t)-\boldsymbol{x}(t)$ is \textit{approximately} Gaussian, whose mean and variance are determined by the Euler-Maruyama approximation  \citep{maruyama1955continuous} (an extension of the Euler method for ordinary differential equations)
\begin{equation*}
\boldsymbol{x}(t+\Delta t) \approx \boldsymbol{x}(t) + g_{\theta}(\boldsymbol{x}(t)) \Delta t + \sqrt{\Delta t} \ \boldsymbol{\epsilon}_t, 
\end{equation*}
with $\boldsymbol{\epsilon}_t\sim \mathcal{N}(0,I)$; $g_{\theta}(\cdot)$ is a function parameterized by $\theta\in\mathbb{R}^m$. 
Assuming that the continuous process is observed \textit{without error} at discrete times $0=t_0<t_1<\cdots<t_n$, it gives rise to an \textit{observed trajectory} of the underlying process
$\{\boldsymbol{x}_{t_i}\}_{i=0}^n$. The log-likelihood of the parameter $\theta$ given the data is given by
\begin{align*}
L(\theta; \{\boldsymbol{x}_{t_i}\}_{i=0}^n)=
\sum_{i=1}^n \ell_i(\theta),\qquad\text{with}~~&\ell_i:=\log\big(\mathcal{K}_{\Delta t_i} (\boldsymbol{x}_{t_i}, \boldsymbol{x}_{t_{i-1}}; \theta)\big),
\end{align*}
wherein $\mathcal{K}_{\Delta t_i}(\cdot,\cdot;\theta)$ denotes the transition kernel for two consecutive observations. Note that $\mathcal{K}_{\Delta t_i}$ corresponds to the discrete time analogue for the observation process of the $\pi_{t|s}$ kernel for the state process discussed in the context of the state identification problem. 

However, in almost all cases, this transition kernel and thus its log-likelihood are analytically intractable. Hence, obtaining the maximum likelihood estimate of $\theta$ is particularly challenging, even though its theoretical properties under ergodicity assumptions are well established \citep{gobet2002lan}.
A number of directions for computing estimates of $\theta$ have been investigated in the literature. 
One direction focuses on the method of moments, or solving estimating equations \citep{gourieroux1993indirect,gallant1997estimating}. Another direction considers closed form asymptotic expansions to the transition kernels $\mathcal{K}_{\Delta t_i}$
\citep{ait2004disentangling,chang2011approximate}. A third direction aims to augment the observed trajectories through imputation at additional time points, so that a satisfactory complete-data approximation of the likelihood function can be written down and subsequently use Markov Chain Monte Carlo methods to impute the additional data and also obtain samples from the posterior distribution of the parameter $\theta$ \citep{elerian2001likelihood,eraker2001mcmc,roberts2001inference}. Additional details on these three broad directions are available in the review paper by \cite{sorensen2004parametric}. Another line of work leverages Monte Carlo techniques for exact simulation of sample paths from the diffusion model \citep{beskos2005exact} and then obtain a Monte Carlo estimate of $\theta$ \citep{beskos2006exact,blanchet2020exact,craigmile2023statistical}.